\documentclass{article} 
\usepackage{iclr2022_conference,times}

\usepackage{amsmath,amsfonts,bm}

\def\eqref#1{equation~\ref{#1}}

\def\1{\bm{1}}

\DeclareMathAlphabet{\mathsfit}{\encodingdefault}{\sfdefault}{m}{sl}
\SetMathAlphabet{\mathsfit}{bold}{\encodingdefault}{\sfdefault}{bx}{n}

\iclrfinalcopy

\usepackage{natbib,hyperref}
\usepackage{url}

\usepackage[utf8]{inputenc} 
\usepackage[T1]{fontenc}    
\usepackage{hyperref}       
\usepackage{url}            
\usepackage{booktabs}       
\usepackage{amsfonts}       
\usepackage{nicefrac}       
\usepackage{microtype}      
\usepackage{amsmath}
\usepackage{graphicx}
\usepackage{graphicx} 
\usepackage{subfigure}
\usepackage{booktabs}
\usepackage{graphicx}
\usepackage{tabularx}
\usepackage{epsfig}
\usepackage{cite}
\usepackage{ntheorem}
\usepackage{diagbox}
\usepackage{rotating}
\usepackage{balance}
\usepackage{enumitem}

\usepackage{hhline}
\usepackage{multirow}
\usepackage{textcomp}
\usepackage{verbatim}
\usepackage{color}
\usepackage{caption}
\usepackage{threeparttable}
\usepackage{ntheorem}
\usepackage{slashbox}
\usepackage{rotating}
\usepackage{algorithm}
\usepackage{algpseudocode}
\usepackage[font=small,labelfont=bf,tableposition=top]{caption}

\DeclareCaptionLabelFormat{andtable}{#1~#2  \&  \tablename~\thetable}

\usepackage{xspace}
\newcommand{\prjname}{DKM\xspace}

\title{\prjname:  Differentiable $k$-Means Clustering Layer  for Neural Network Compression }

\author{
 Minsik Cho\thanks{equal contribution} \\
  \And
 Keivan Alizadeh-Vahid\footnotemark[1] \\
  \And
 Saurabh Adya \\
  \And
  Mohammad Rastegari \\
  \And \hspace{2.45in}Apple\\\\
   \hspace{0.4in} \texttt{\{minsik, kalizadehvahid, sadya, mrastegari\}@apple.com} 
} 

\begin{document}

\maketitle

\begin{abstract}
Deep neural network (DNN) model compression for efficient on-device inference becomes  increasingly important to reduce memory requirements and keep user data on-device.
To this end, we propose
 a novel differentiable $k$-means clustering layer (\prjname) and its application to train-time weight-clustering for DNN model compression.
\prjname casts   $k$-means clustering as an attention problem and enables
joint  optimization of the DNN parameters and clustering centroids.
Unlike prior works that rely on additional parameters and regularizers,
\prjname-based compression keeps the original loss function and model architecture fixed.
We evaluated  \prjname-based compression on various DNN models  for computer vision and natural language processing (NLP) tasks. Our results demonstrate that \prjname delivers superior compression and accuracy trade-off on ImageNet1k and GLUE benchmarks. For example,    \prjname-based compression can offer 74.5\% top-1 ImageNet1k accuracy on ResNet50   with 3.3MB model size (\textbf{29.4}x model compression factor). 
For MobileNet-v1, which is a challenging DNN to
compress, \prjname delivers 63.9\% top-1 ImageNet1k accuracy with 0.72 MB model size (\textbf{22.4x} model compression factor). This result is 6.8\% higher top-1 accuracy and 33\% relatively smaller model size than the current state-of-the-art DNN compression algorithms.  
 \prjname also compressed a DistilBERT model by \textbf{11.8x} with minimal (1.1\%) accuracy loss on GLUE NLP benchmarks.

\end{abstract}

\section{Introduction}
Deep neural networks (DNN) have demonstrated super-human 
performance on many cognitive tasks~\citep{SilHub18General}. While a fully-trained uncompressed DNN is commonly used for server-side inference, on-device inference is preferred to enhance user experience by reducing latency and keeping user data on-device.
Many such on-device platforms are battery-powered and resource-constrained, demanding
a DNN to meet the stringent resource requirements such as power-consumption, compute budget and storage-overhead~\citep{haq,pmlr-v80-wu18h}.

One solution is to design a more efficient and compact DNN such as MobileNet~\citep{mobilenet} by innovating the network architecture or by leveraging Neural Architecture Search (NAS) methods ~\citep{liu2019darts, tan2019mnasnet}.
Another solution is to compress a model with small accuracy degradation so that it takes less storage and reduces System-on-Chip (SoC) memory bandwidth utilization, which can minimize power-consumption and latency. 
To this end, various DNN compression techniques have been proposed~\citep{haq, hawq-v2, Park_2018_ECCV,xor_net,quant_noise,stock2020bit, zhou2019neural, park2019lookahead, yu2018nisp, polino2018model}. Among them, weight-clustering/sharing~\citep{deepcomp_iclr16,pmlr-v80-wu18h, ullrich2017soft,stock2020bit} has shown a high DNN compression ratio where weights are clustered  into a few shareable weight values (or centroids) based on 
$k$-means clustering. Once weights are clustered, to shrink the model size, one can store indices (2bits, 4bits, etc. depending on
the number of clusters) with a lookup table rather than actual floating-point values.

 Designing a compact DNN architecture and enabling weight-clustering together could provide the best solution in terms of efficient on-device inference. However, the existing model compression approaches do not usefully compress an already-compact DNN like MobileNet, presumably because the model itself does not have significant redundancy. 
 We conjecture that such limitation comes from the fact that  weight-clustering through $k$-means algorithm (both weight-cluster assignment and weight update) has not been fully optimized with the target task. The fundamental complexity in applying $k$-means clustering for weight-sharing comes from the following: \textbf{a)} both weights and corresponding k-means centroids are free to move (a general $k$-means clustering with fixed observations is already NP-Hard), \textbf{b)} the weight-to-cluster assignment is a discrete process which makes $k$-means clustering non-differentiable, preventing effective optimization.

In this work, we propose a new layer without learnable parameters for differentiable $k$-means clustering, \prjname, based on an attention mechanism~\citep{attention_iclr15} to capture the weight and cluster interactions seamlessly,  and further apply it to enable train-time weight-clustering for model compression. Our major contributions include the following:

\begin{itemize}
\vspace{-0.1 in}
    \item We propose a novel differentiable $k$-means clustering layer (\prjname) for deep learning, which serves as a generic neural layer to develop clustering behavior on input and output.
        \item We demonstrate that \prjname can perform multi-dimensional $k$-means clustering efficiently and can offer a high-quality model for a given compression ratio target.
    \item We apply \prjname to compress a DNN model and demonstrate the state-of-the-art results on both computer vision and natural language models and tasks.
\end{itemize}


\section{Related Works}

\label{prior:diffkmeans}

\textbf{Model compression using clustering}: 
DeepCompression~\citep{deepcomp_iclr16} proposed to apply $k$-means clustering for model
compression. 
DeepCompression initially clusters the weights using $k$-means algorithm. All the weights that belong to the same cluster share the same weight value which is initially the cluster centroid. In the forward-pass, the shared weight is used for each weight. In the backward-pass, the gradient for each shared weight is calculated and used to update the shared value.
This approach might degrade model quality because it cannot formulate weight-cluster assignment during gradient back propagation~\citep{yin2019understanding}.
ESCQ~\citep{limit_quant, univ_deep_comp} is optimizing the clusters to minimize the change in the loss by considering hessian. Therefore, it is to preserve the current model state, instead of searching for a fundamentally better model state for compression.


HAQ~\citep{haq} uses reinforcement learning to search for the optimal quantization policy on different tasks. For model compression, HAQ uses $k$-means clustering similar to DeepCompression yet with flexible bit-width on different layers. Our work is orthogonal to this work because the $k$-means clustering can be replaced with our \prjname with a similar flexible configuration. "And The Bit Goes Down"~\citep{stock2020bit} algorithm is based on Product Quantization and Knowledge Distillation. It evenly splits the weight vector of $N$ elements into $N/d$ contiguous $d$ dimensional sub-vectors, and clusters the sub-vectors using   weighted $k$-means clustering to minimize activation change from that of a teacher network. GOBO \citep{gobo} first separates outlier weights  far from the average of the weights of each layer  and stores them uncompressed while clustering the other weights   by an algorithm similar to $k$-means.

\textbf{Model compression using regularization}: 
Directly incorporating $k$-means clustering in the training process is not straightforward~\citep{pmlr-v80-wu18h}. Hence, \citep{ullrich2017soft} models weight-clustering as Gaussian Mixture Model (GMM) and fits weight distribution into GMM with additional learning parameters using KL divergence (i.e., forcing weight distribution to follow $k$ Gaussian distributions with a slight variance).
\citep{pmlr-v80-wu18h} proposed deep $k$-means to enable weight-clustering during re-training. By forcing the weights that have been already clustered to stay around the assigned center, the hard weight-clustering is approximated with additional parameters.
Both  \citep{ullrich2017soft} and \citep{pmlr-v80-wu18h} leverage regularization to enforce weight-clustering with additional parameters, which will interfere with the original loss target and requires additional updates for the new variables (i.e., singular value decomposition (SVD) in \citep{pmlr-v80-wu18h}). Also, relying on the modified loss cannot capture the dynamic interaction between weight distributions and cluster centroids within a batch, thus requiring an additional training flow for re-training.

\textbf{Enhance Model compression using dropout}: 
Quant-Noise~\citep{quant_noise} is  a structured dropout which only quantizes a random subset of weights (using any quantization technique) and thus can improve the predictive power of a compressed model. For example, \citep{quant_noise} showed good compression vs. accuracy trade-off on ResNet50 for ImageNet1k.

\textbf{Model quantization}: Besides clustering and regularization methods, model quantization can also reduce the model size, and training-time quantization techniques have been developed to improve the accuracy of quantized models ~\citep{li2019additive, zhao2019linear}. EWGS~\citep{lee2021network} adjusts gradients by scaling them up or down based on the Hessian approximation for each layer.
PROFIT \citep{park2020profit} adopts an iterative process and freezes layers based on the activation instability.

\textbf{Efficient networks}: Memory-efficient DNNs include MobileNet \citep{mobilenet, mobilenet_v2}, EfficientNet \citep{tan2020efficientnet, tan2021efficientnetv2} and ESPNet \citep{mehta2019espnetv2}.  MobileNet-v1 \citep{mobilenet}  on ImageNet1k dataset has top-1 accuracy of 70.3\% with 16.1 MB of memory in comparison to a ResNet18 which has 69.3\% accuracy with 44.6 MB of model size. Our method can be applied to these compact networks to  reduce their model sizes further.


\section{Algorithm}

\label{body:diffkmeans}

\begin{figure}[!b]
	\begin{center}
		\mbox{
			\subfigure[Conventional weight-clustering~\citep{deepcomp_iclr16,haq,stock2020bit,lee2021network}]
			{\includegraphics[width=2.7in]{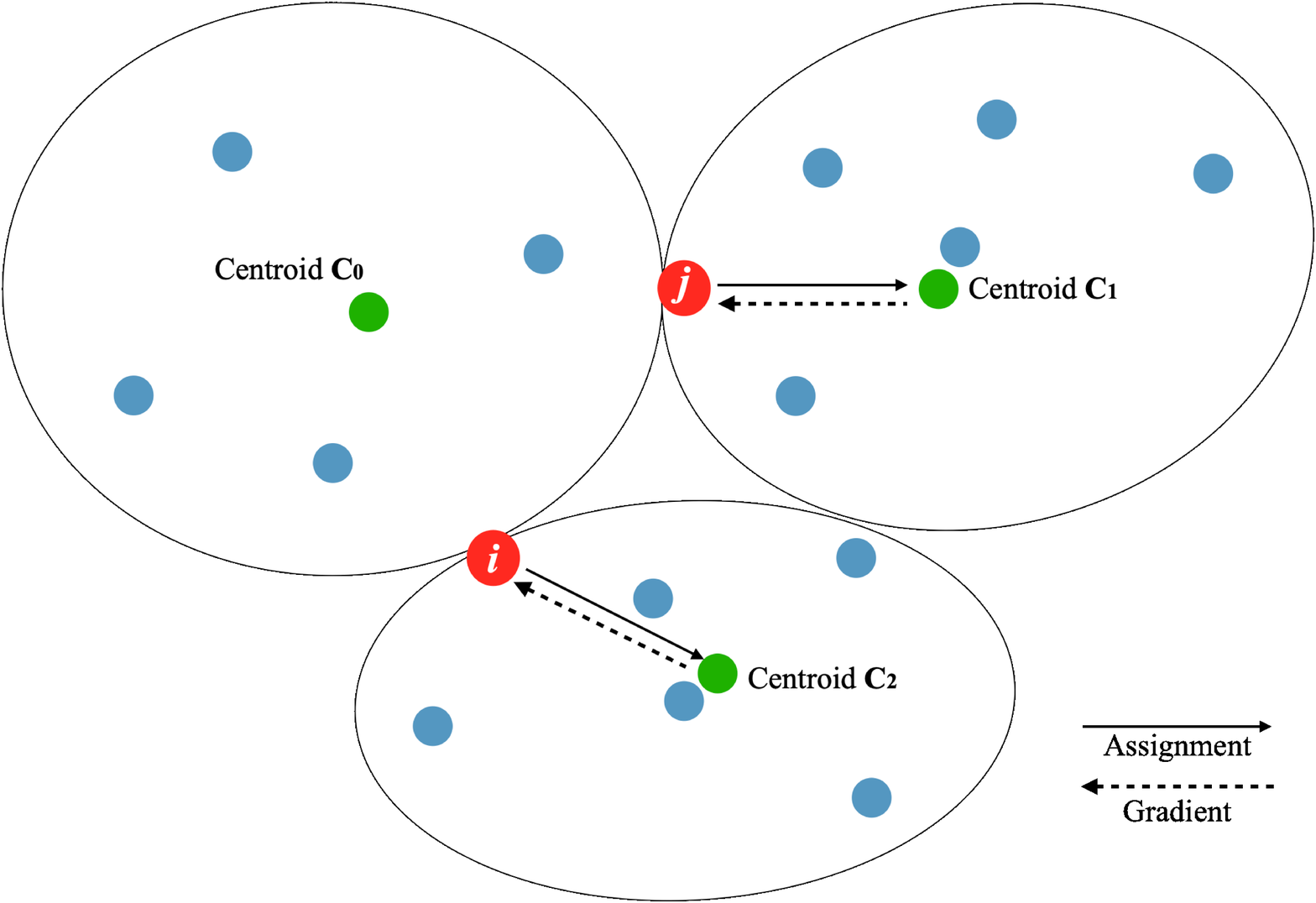}} 

			\subfigure[Attention-based weight-clustering in \prjname]
			{\includegraphics[width=2.7in]{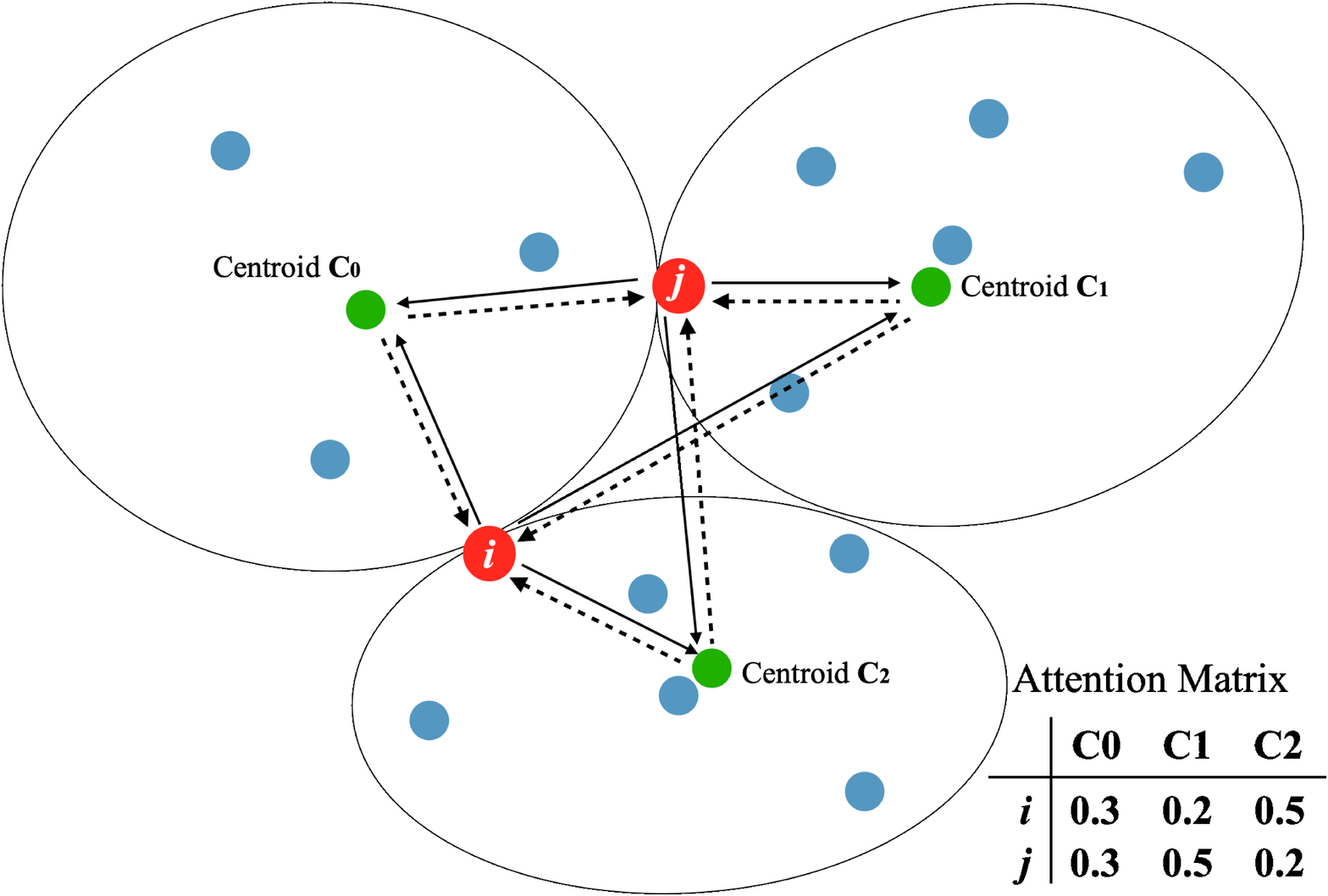}} \quad
		}
		\caption{In conventional weight-clustering algorithms, the boundary weights $i $ and $j$ are assigned to clusters $C_2$ and $C_1$ based on the distance metric respectively, which is neither necessarily suitable for the task nor differentiable against the loss function.  \prjname instead applies soft assignment using attention mechanism during forward-propagation and enables differentiable backward-propagation, which 
		allows weights to consider other non-nearest clusters (especially helpful for the boundary weights) and shuttle
		among multiple clusters in order to directly optimize their assignments based on the task loss function.
		}
		\label{comp_tdk_vs_others}
	\end{center} 
\end{figure}


\subsection{Motivation}
\label{body:overview}
Popular weight-clustering techniques for DNN model compression~\citep{lee2021network,deepcomp_iclr16, hawq-v2,stock2020bit} are based on $k$-means clustering along with enhancements such as
gradient scaling/approximation. Using $k$-means clustering, the weights are clustered and assigned to the nearest centroids which are used for forward/backward-propagation during training as illustrated in Fig.~\ref{comp_tdk_vs_others} (a).  Such conventional methods with clustering have two critical drawbacks:
\begin{itemize}
    \item The weight-to-cluster assignment in conventional approaches is not optimized through back-propagation of training loss function. 
    \item Gradients for the weights are computed in an ad-hoc fashion: the gradient of a centroid is re-purposed as the gradient of the weights assigned to the centroid.
\end{itemize}    
These limitations are more pronounced for the weights on the boundary such as $i$ and $j$ in Fig.~\ref{comp_tdk_vs_others} (a).
In the conventional approaches, $i$ and $j$ are assigned to the centroids $C_2$ and $C_1$ respectively, simply because of
their marginal difference in a distance metric. However, assigning $i$ to $C_0$ and $j$ to $C_2$ could be  better for the training loss as their difference in distance is so small~\citep{pmlr-v119-nagel20a}. Such lost opportunity cost is especially higher with a smaller number of centroids (or fewer bits for quantization), as each \textit{unfortunate} hard assignment can  degrade the training loss significantly.

\begin{figure}[t]
  \centering
  \includegraphics[width=5 in]{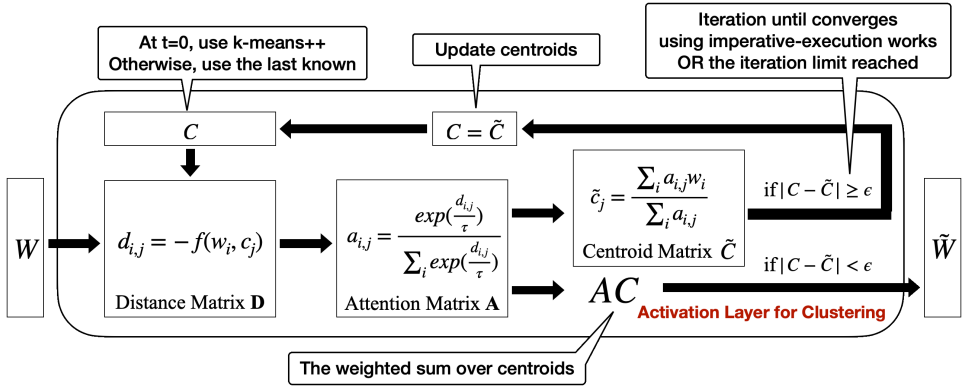}
  \caption{ Weight-sharing using attention matrix $\mathbf{A}$ is iteratively performed in a \prjname layer until the centroids ($C$) converge. Once converged, a compressed weight, $\tilde W$ is used for  forward-propagation. Since \prjname is a differentiable layer, backward-propagation will run through the iterative loop and the gradients for the weights will be computed against the task loss function.}
  
  \label{fig:architecture}
\end{figure}

We overcome such limitations with \prjname by interpreting  weight-centroid assignment  as distance-based attention optimization~\citep{attention_iclr15} as in
Fig.~\ref{comp_tdk_vs_others} (b) and letting each weight interact with all the centroids. Such attention mechanism naturally cast differentiable and iterative $k$-means clustering into a parameter-free layer  as in Fig.~\ref{fig:architecture}.
Therefore, during backward-propagation,   attention allows a gradient of a weight to be a product of the attentions and the gradients of centroids, which in turn impact how the clustering and assignment will be done in the next batch. overall our weight assignment will align with the loss function, and can be highly effective for DNN compression.

\subsection{Differentiable K-means clustering layer for Weight-clustering}
\label{body:core}

\prjname can perform a differentiable train-time weight-clustering iteratively for $k$ clusters as shown in Fig.~\ref{fig:architecture} for the DNN model compression purpose. Let $\mathbf{C} \in R^{k}$ be a vector of cluster centers and $\mathbf{W} \in R^{N}$ be a vector of the weights, and then \prjname performs as follows:

\begin{itemize}
\vspace{-0.08 in}
    \item In the first iteration $\mathbf{C}$ can be  initialized either by randomly selected $k$ weights from  $\mathbf{W}$  or   $k$-means++. For all subsequent iterations, the last known $\mathbf{C}$ from the previous batch is used to  accelerate the clustering convergence.
    
    \item A distance matrix $\mathbf{D}$ is computed for every pair between a weight $w_i$ and
    a centroid $c_j$ using a differentiable metric $f$ (i.e., the Euclidean distance) using $d_{ij} = -f(w_i, c_j)$.
    
    \item We apply softmax with a temperature $\tau$ on each row of $\mathbf{D}$ to obtain attention matrix $\mathbf{A}$ where $a_{i,j} = \frac{exp(\frac{d_{i,j}}{\tau})}{\sum_{k}exp(\frac{d_{i,k}}{\tau})}$  represents
    the attention from $w_i$ and $c_j$.
    
    \item Then, we obtain a centroid candidate, $\mathbf{\tilde C}$ by gathering all the attentions from $\mathbf{W}$ for each centroid by computing
    $\tilde c_j = \frac{\sum_{i} {a}_{i,j} {w}_i}{\sum_{i} {a}_{i,j}}$
    and update $\mathbf{ C}$ with $\mathbf{\tilde C}$ if the iteration continues.

    \item We repeat this process till $|\mathbf{C}-\mathbf{\tilde C}| \le \epsilon$  at which point $k$-means has converged or the iteration limit reached, and we compute $\mathbf{A}\mathbf{C}$ to get 
     $\mathbf{\tilde W}$ for forward-propagation (as in Fig.~\ref{fig:architecture}). 
     
\end{itemize}

The iterative process will be dynamically executed imperatively in PyTorch~\citep{pytorch_eager} and Tensorflow-Eager~\citep{tf_eager}  and is differentiable for backward-propagation, as $\mathbf{\tilde W}$ is based on the
attention between weights and centroids. \prjname uses soft weight-cluster assignment which could be hardened in order to impose weight-clustering constraints. The level of hardness can be controlled
by the temperature $\tau$ in the softmax operation.  
During inference we 
use  the last attention matrix (i.e., $\mathbf{A}$ in Fig.~\ref{fig:architecture}) from a \prjname layer to snap each weight to the closest centroid of the layer and finalize weight-clustering as in prior arts (i.e., no more attention), but such assignment is expected to be tightly aligned with the loss function, as the weights have been annealed by shuttling among centroids. A theoretical interpretation of DKM is described in Appendix \ref{appendix:em}. 

Using \prjname for model compression allows a weight to change its cluster assignment during training, but eventually encourages it to settle with the best one w.r.t the task loss. Optimizing both weights and clusters simultaneously and channeling the loss directly to the weight-cluster assignment is by the attention mechanism. Since \prjname is without additional learnable parameters and  transparent to a model and loss function, we can reuse the existing training flow and hyper-parameters. 
The key differences between \prjname-based compression and the prior works can be summarized as follows:

\begin{itemize}
    \item Instead of hard weight-cluster assignment and approximated gradient~\citep{deepcomp_iclr16, haq, lee2021network, stock2020bit}, \prjname uses flexible and differentiable attention-based weight-clustering and computes gradients w.r.t the
    task loss.
    
    \item Instead of modifying the loss function with regularizers to enforce clustering~\citep{ullrich2017soft,pmlr-v80-wu18h},
    \prjname  can be inserted   into forward pass, making the optimization fully aligned with the task objective (i.e., no interference in loss).
    
    \item \prjname requires  no additional learnable parameters~\citep{ullrich2017soft,lee2021network}, thus making the training flow simple. For example, \prjname-base approach does not need to substitute a convolution layer with a specialized version with additional learning parameters. 
    
    \item \prjname requires no additional computation such as Hessian trace~\citep{hawq-v2,limit_quant} or SVD~\citep{lee2021network,pmlr-v80-wu18h} for gradient approximation, because \prjname uses a differentiable process.
    
    \item \prjname-based compression does not require  a complex training techniques such as freezing/progress~\citep{park2020profit} or distillation~\citep{stock2020bit}.
    
\end{itemize}

  \begin{figure}[!b]
\setlength{\tabcolsep}{8 pt}
\renewcommand{\arraystretch}{1.6}  
    \centering
    \includegraphics[width=1.4in]{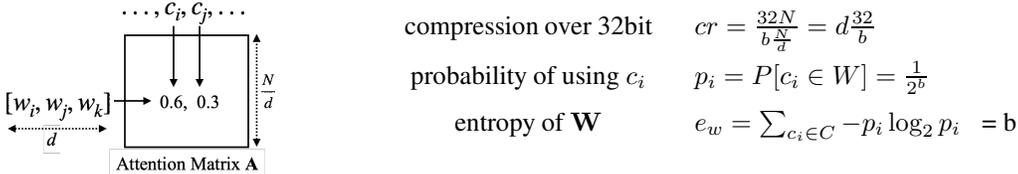}
    \hspace{0.5 in}
    \begin{tabular}[b]{cc}
      compression over 32bit &  \multicolumn{1}{l}{$cr=\frac{32N}{b\frac{N}{d}} = d\frac{32}{b}$ }\\
      probability of using $c_i$ & \multicolumn{1}{l}{$p_i = P[c_i \in W] = \frac{1}{2^b}$}\\
      entropy of $\mathbf{W} $ & \multicolumn{1}{l}{$ e_w = \sum_{c_i \in C} -p_i \log_2  p_i$} = b\\
      &
    \end{tabular}
    \caption{Multi-dimensional \prjname can increase the weight entropy ($e_w$) with a fixed compression ratio  target ($cr$) by increasing a dimension ($d$), which may improve the model accuracy~\citep{weight_entropy}.}
    \label{multiv}
  \end{figure}
  
\subsection{Multi-Dimensional \prjname}
\label{body:multid}
\prjname can be naturally extended into multi-dimensional weight-clustering~\citep{stock2020bit} due to 
its simplicity, and is highly effective due to its differentiability.
We split $N$ elements of weights into $\frac{N}{d}$ contiguous $d$ dimensional sub-vectors and cluster the sub-vectors ($\mathbf{W} \in R^{\frac{N}{d}*d}$). 
For example, we simply flatten all the convolutional kernels into a $(\frac{N}{d},d)$ matrix across both kernel and channel boundaries and apply multi-dimensional \prjname to the matrix for clustering in our implementation as in Fig.~\ref{multiv}.
Accordingly, the cluster centroids will become $d$-dimensional as well ($\mathbf{C} \in R^{k*d}$) and the metric calculation is done in the $d$-dimensional space. 
With the multi-dimensional scheme, the effective \textit{bit-per-weight} becomes $\frac{b}{d}$ for   $b$-bit/$d$-dim clustering. The memory complexity of a \prjname layer with $N$ parameters is  $O(r\frac{N}{d}2^b)$ where $r$ is the number of iterations per Fig.~\ref{fig:architecture} (i.e., all the intermediate results such as $\mathbf{D}$ and $\mathbf{A}$ at each iteration need to be kept for backward-propagation).

Such multi-dimensional clustering
could be ineffective for conventional methods (i.e., DNN training not converging)~\citep{stock2020bit,haq,lee2021network}, as now a weight might be on the boundary to multiple centroids, and the chance of making wrong decisions grows exponentially with the number of centroids. For example, there are only two centroids for 1bit/1dim clustering, while there are 16 centroids in 4bit/4dim clustering, although both have the same effective \textit{bit-per-weight}. Intuitively, however, \prjname can work well with such multi-dimensional configurations as \prjname  naturally optimizes the assignment w.r.t the task objective and  can even recover from a wrong assignment decision over the training-time optimization process.

The key benefit of multi-dimensional \prjname is captured in Fig.~\ref{multiv}. For a given  $N \gg d$ in 32 bits,
the compression ratio is $d\frac{32}{b}$ (i.e., a function of both $d$ and $b$). Assuming the number of sub-vectors assigned to each centroid is same, the entropy of $\mathbf{W}$
is simply $b$.  Since higher entropy  in the weight distribution indicates larger learning capacity and better model quality~\citep{weight_entropy},   increasing $b$ and $d$ at the same ratio as much as possible may improve the model quality for  a given target compression ratio (see Section~\ref{result:imagenet1k} for results). However, making $b$ and $d$ extremely large to maximize the entropy of $\mathbf{W}$ might be impractical, as the memory overhead grows exponentially with $b$.

\section{Experimental Results}
  \begin{figure}[!t]
\setlength{\tabcolsep}{8 pt}
\renewcommand{\arraystretch}{0.95}  
    \centering
	\mbox{
			\subfigure[ResNet18 ]
			{\includegraphics[width=2.8in]{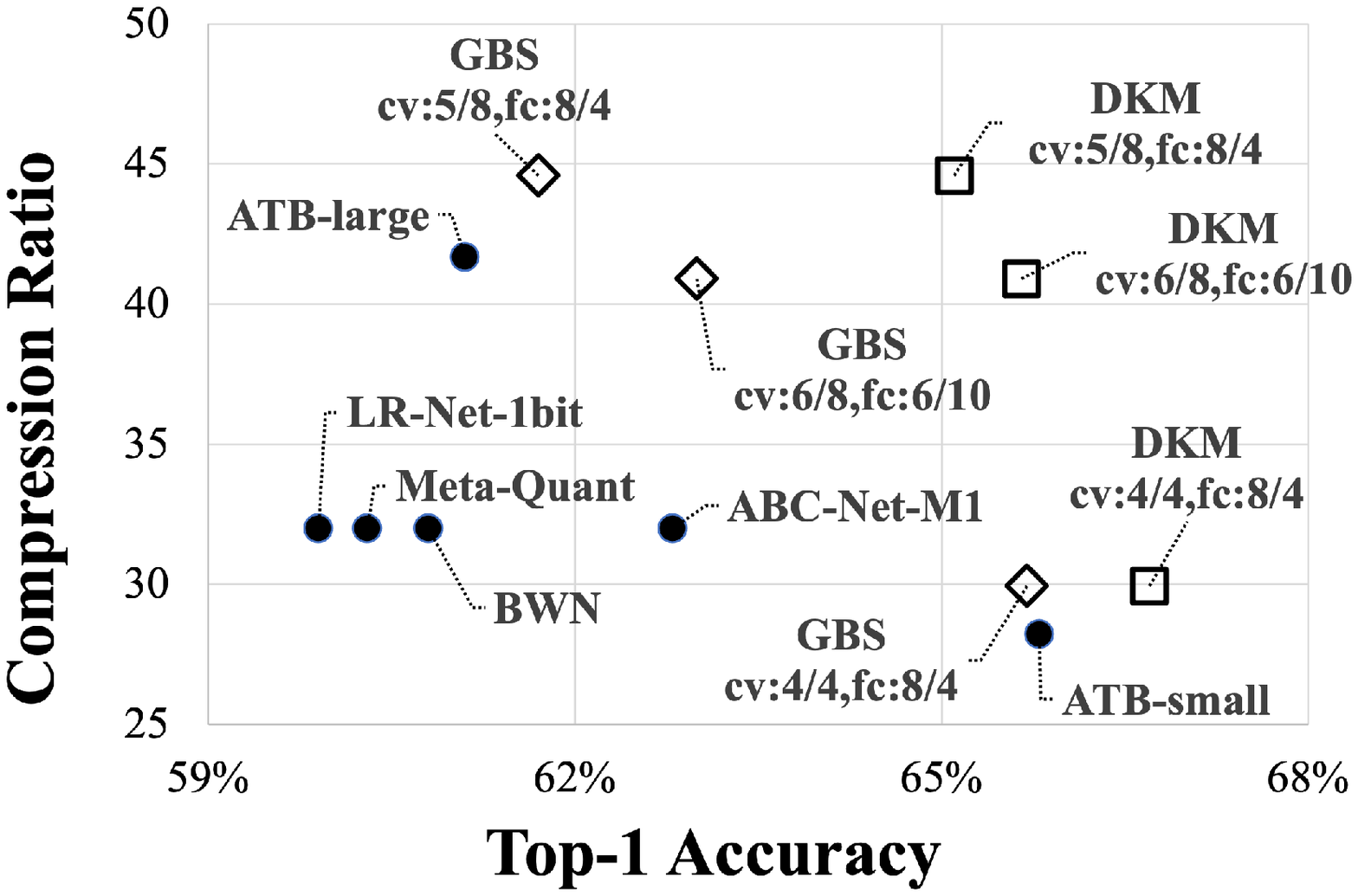}} 

			\subfigure[ResNet50  ]
			{\includegraphics[width=2.8in]{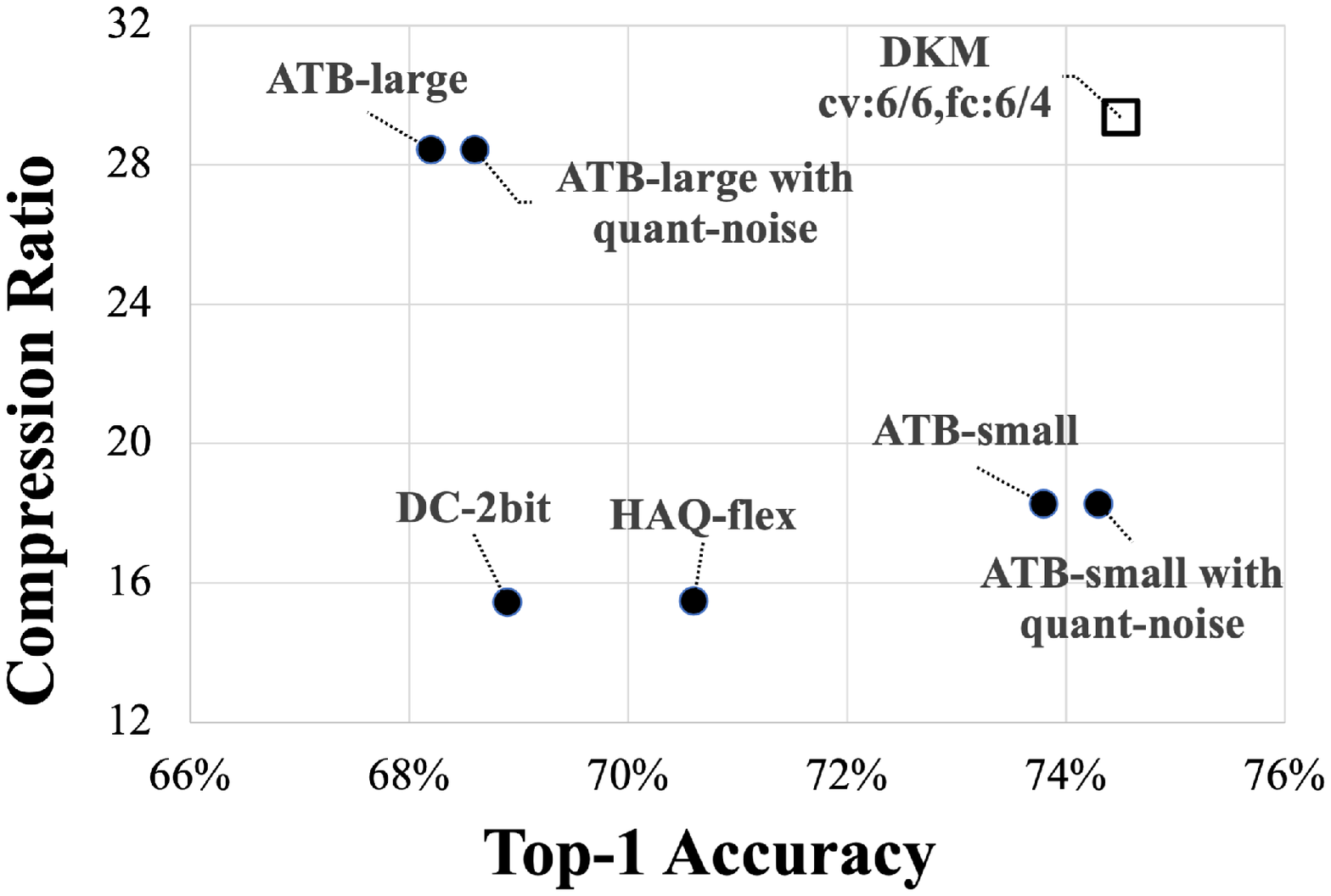}}



		}
\begin{threeparttable}
\setlength{\tabcolsep}{5 pt}
\renewcommand{\arraystretch}{0.95}
\small
\begin{tabular}{cc|c|cccc|ccccccc}
\\\hline
\multicolumn{2}{c|}{ }   &  Base      & DC   & HAQ  & ATB  & ATB  & \multicolumn{3}{c}{\prjname}
\\
   
   Model&Metrics & 32bit &2bit & flex & small & large & \multicolumn{2}{c}{configuration} & b/w$^*$\\
   \hline\hline   
   
   
    \multirow{2}{*}{ResNet18} & Top-1 (\%)   & 69.8  &       &       & 65.8 & 61.1  & {65.1}$^{	\checkmark}$ & cv$^\dagger$:5/8$^\mathsection$& \multirow{2}{*}{0.717}\\   
                              & Size (MB)    & 44.6  &       &       & 1.58 & 1.07  & \textbf{1.00} & fc$^\ddagger$:$8/4$& \\   \hline
    
    \multirow{3}{*}{ResNet50} &  \multirow{2}{*}{Top-1 (\%)}   & \multirow{2}{*}{76.1}  & \multirow{2}{*}{68.9}  & \multirow{2}{*}{70.6}  & 73.8  & 68.2 & \multirow{2}{*}{\textbf{74.5}} & \multirow{2}{*}{cv:$6/6$} &  \multirow{3}{*}{1.077}  \\  
                              &         &       &       &       & 74.3$^{\triangle}$  & 68.8$^{\triangle}$ &  \\  
                              & Size (MB)    & 97.5  & 6.32  & 6.30  & 5.34 & 3.43 & \textbf{3.32} & fc:$6/4$ & \\    \hline

    MobileNet                 & Top-1 (\%)  & 70.9  & 37.6  & 57.1  & \multirow{2}{*}{nc$^\circ$} &\multirow{2}{*}{nc} & \textbf{63.9} & cv:$4/4$ &  \multirow{2}{*}{1.427} \\  
    v1                        & Size (MB)    & 16.1  & 1.09  & 1.09  &                      &                   & \textbf{0.72} & fc:$4/2$ &\\\hline   
                   
    MobileNet                 & Top-1 (\%)  & 71.9  & 58.1  & 66.8  & \multirow{2}{*}{nc} &\multirow{2}{*}{nc} & \textbf{68.0} & cv:$2/1$ & \multirow{2}{*}{2.010} \\  
        v2                    & Size (MB)    & 13.3  & 0.96  & 0.95  &                     &                    & \textbf{0.84} & fc:$4/4$ &\\\hline

\end{tabular} 
\begin{tablenotes}
\item[] $^*$ effective bit-per-weight (see Section~\ref{body:multid}); $^\circ$ not converging
\item[]  $^\dagger$ the convolution layers ; $^\ddagger$ the last fully connected layer;
\item[] $^\mathsection$ clustering with $6$ bits and $8$ dimensions; $^{\triangle}$ ATB with quantization-noise~\citep{quant_noise}
\item[] $^{	\checkmark}$ also, 65.7\% Top-1 accuracy and 1.09 MB with cv:6/8, fc:6/10 
\item[] \quad \quad \quad 66.7\% Top-1 accuracy and 1.49 MB with cv:4/4, fc:8/4 
\end{tablenotes}

\end{threeparttable}
\caption{  Accuracy vs. compression trade-offs: \prjname-powered compression (in white box markers) delivers the Pareto superiority to the other schemes (i.e., the top-right corner is the best trade-off) for ResNet18, ResNet50, and MobileNet-v1/v2 on ImageNet1k. }
\label{result_size_imagenet1k}
\vspace{-0.1 in}
  \end{figure}

\label{result:diffkmeans}
We compared our \prjname-based compression with state-of-the-art quantization or compression schemes on various computer vision and natural language models. 
To study the trade-off between model compression and accuracy, we disabled activation quantization in every experiment for all approaches, as our main goal is the model   compression.
All our experiments with \prjname were done on two x86 Linux machine with eight NVIDIA V100 GPUs each in a public cloud infrastructure. We used a SGD optimizer with momentum 0.9, and fixed the learning rate at 0.008 (without individual hyper-parameter tuning) for all the experiments for \prjname. Each compression scheme starts with publicly available pre-trained models. The $\epsilon$ is set as $1\text{e--}4$ and the iteration limit is 5.

\begin{table}[t]
\vspace{-0.2 in}
\centering
\small
\begin{threeparttable}
\renewcommand{\arraystretch}{0.95}

\begin{tabular}{cc|cccccc}
\hline
   \multicolumn{2}{c|}{}   & ResNet18  & ResNet50  & MobileNet-v1 & MobileNet-v2  \\  
    &    Base (32 bit)            & 69.8     & 76.1     & 70.9    &  71.9  \\  \hline \hline
    
    \multirow{4}{*}{{\rotatebox[origin=c]{90}{3 bit}}} 
    &PROFIT     &  &       & 69.6         &  69.6    \\  
    &EWGS     & \textbf{70.5} & \textbf{76.3} & 64.4 & 64.5 \\   
    &PROFIT+EWGS                     & &   & 68.6         &  69.5  \\  
    &\prjname                     & 69.9      & 76.2     & \textbf{69.9}   & \textbf{70.3} \\  \hline \hline
    
    \multirow{3}{*}{{\rotatebox[origin=c]{90}{2 bit}}} 
    & PROFIT   &            &          &    63.4     &   61.9 \\   
    & EWGS     &   \textbf{69.3} & \textbf{75.8} &    52.0   &  49.1 \\  
    & \prjname     & 68.9     & 75.3     & \textbf{66.4}         &  \textbf{66.2} \\  \hline \hline

    \multirow{5}{*}{{\rotatebox[origin=c]{90}{1 bit}}} 
    & PROFIT    &          &         &    nc$^\circ$       &   nc \\   
    & EWGS      & 66.6 &  73.8 & 8.5 &  23.0   \\ 
    &\prjname 1/1      & 65.0     & 72.1    & 5.9         & 50.8 \\  
    &\prjname  $4/4^\mathsection$          & 67.0&  \textbf{73.8} &  60.6  &  55.0  \\ 
    &\prjname   $8/8$  &  \textbf{67.8} &oom$^\Box$ & \textbf{64.3} & \textbf{62.4}    \\   \hline 
    \hline
    
    \multirow{2}{*}{{\rotatebox[origin=c]{90}{$\frac{1}{2}$ bit}}} 
    &\prjname   $4/8$  & 62.1     &70.6    & 46.5   & 34.0         \\   
    &\prjname   $8/16$ & \textbf{65.5} & \textbf{72.1} & \textbf{59.8} & \textbf{58.3}  \\  \hline
   
\end{tabular}

\begin{tablenotes}
\item[] $^\circ$ not converging; $^\Box$ out of memory ; $^\mathsection$ clustering with $4$ bits and $4$ dimensions
\end{tablenotes}

\end{threeparttable}
 \vspace{0.1 in}

\caption{When compared with the latest weight quantization algorithms, \prjname-based algorithm shows superior Top-1 accuracy when the network is hard to optimize (i.e., MobileNet-v1/v2) or when a low precision is required (1 bit). Further, with multi-dimensional \prjname (see Section~\ref{body:multid}), \prjname delivers
64.3 \% Top-1 accuracy for MobileNet-v1 with the 8/8 configuration which is equivalent to 1 bit-per-weight.}
\label{result_top1_imagenet1k}
\vspace{-0.2 in }
\end{table}

\begin{table}[t]
\centering
\small
\begin{threeparttable}
\setlength{\tabcolsep}{5 pt}
\renewcommand{\arraystretch}{0.95}

\begin{tabular}{cc|cccccc}
\hline
      Metrics &   Base (32bit)   & RPS     & \prjname 4/1  & \prjname 4/2 & \prjname 8/8$^\mathsection$  \\ 
  \hline\hline   
          Top-1& 69.8         & 67.9  & \textbf{70.9} &  70.3  &  68.5 \\  
                                  cr$^	\Diamond$ &  1   & 4  & 8 & 16 & 32             \\    \hline
 
\hline
\end{tabular} 
\begin{tablenotes}
\item[] $^	\Diamond$ compression ratio; $^\mathsection$ clustering with $8$ bits and $8$ dimensions
\end{tablenotes}
\end{threeparttable}
\vspace{0.1 in}
 
\caption{GoogleNet training performance for ImageNet1k: \prjname-based compression offered 2x better compression ratio with 3\% higher top-1 accuracy than RPS~\citep{pmlr-v80-wu18h}.}
\vspace{-0.2 in}
\label{result_size_googlenet}
\end{table}

\subsection{ImageNet1k}
\label{result:imagenet1k}

We compared our \prjname-based compression with prior arts: DeepCompression (or \textbf{DC})~\citep{deepcomp_iclr16}, \textbf{HAQ}~\citep{haq}, and  "And The Bit Goes Down" (or \textbf{ATB})~\citep{stock2020bit}  combined with Quantization-noise~\citep{quant_noise}, \textbf{ABC-Net}~\citep{abcnet},
\textbf{BWN}~\citep{xor_net}, \textbf{LR-Net}~\citep{lr_net}, and \textbf{Meta-Quant}~\citep{meta_quant}. We also compared with \textbf{GBS} which uses the same flow as \prjname except that \textbf{Gumbel-softmax} is used to generate stochastic soft-assignment as attention~\citep{jang2017categorical}. 
In the \textbf{GBS} implementation, we 
iteratively perform drawing to mitigate the large variance problem reported in~\citep{lr_net}.

We set the mini-batch size 128 per GPU (i.e., global mini-batch size of 2048) and ran for 200 epochs for all  \prjname cases. Since the public ATB implementation does not include MobileNet-v1/v2 cases~\citep{mobilenet,mobilenet_v2}, we added
the support for these two by following the paper and the existing ResNet18/50~\citep{he2016deep} implementations. 
Instead of using a complex RL technique as in  HAQ~\citep{haq}, for \prjname experiments, we fixed
configurations for all the convolution layers (noted as cv) and the last fully connected layer (noted as fc), except that we applied 8 bit clustering to a layer with fewer than 10,000 parameters. 

Using \prjname layer, our compression method offers a Pareto superiority to other schemes as visualized in  Fig.~\ref{result_size_imagenet1k}.
For ResNet50 and MobileNet-v1/v2, \prjname delivered  compression configurations that yielded both better accuracy and higher compression ratio than the prior arts. For ResNet18,  \prjname was able to make a smooth trade-off on accuracy vs. compression, and find Pareto superior configurations to all others:   \prjname can get 65.1\% Top-1 accuracy with 1MB  which is superior to ATB-large,    66.8\% Top-1 accuracy with 1.49 MB which is superior to  ATB-small, and
65.8\% Top-1 accuracy with 1.09 MB as a superior balance point. For MobileNet-v1/v2, ATB failed to converge, but \prjname
outperforms DC and HAQ in terms of both accuracy and size at the same time.
For ATB cases, adding quantization noise improves the model accuracy~\citep{quant_noise} only moderately.
GBS in fact shows better performance than ATB, but still worse than the proposed method, even after reducing variance through iteration, especially for the high compression configurations. For ResNet18, GBS with the same bit/dimension targets delivered the following Top-1 accuracies ranging from 61.7\% to 65.7\%. For details, please refer to Appendix~\ref{appendix:gbs_hard}.



We also compared \prjname with a well-known regularization-based clustering method on GoolgeNet in Table~\ref{result_size_googlenet}, \textbf{RPS}~\citep{pmlr-v80-wu18h}  which has demonstrated superior performance to another regularization approach~\citep{ullrich2017soft}. Note that only convolution layers are compressed, following the setup in RPS~\citep{pmlr-v80-wu18h}. Table~\ref{result_size_googlenet}
clearly indicates that \prjname can allow both much better compression and higher accuracy than RPS even with 1 bit-per-weight.

We also compared our \prjname-based algorithm with the latest scalar weight quantization approaches, \textbf{PROFIT}~\citep{park2020profit} and \textbf{EWGS}~\citep{lee2021network} (which have outperformed the prior arts in the low-precision regimes) by running their public codes on our environments with the recommended hyper-parameter sets. 
Table~\ref{result_top1_imagenet1k} summarizes our comparison results on ResNet18, ResNet50, and MobileNet-v1/v2 for the ImageNet1k classification task. Following the experimental protocol in~\citep{lqnet,lee2021network,xor_net}, we did not compress the first and last layers for all the experiments in Table~\ref{result_top1_imagenet1k}.
It clearly shows that our approach with \prjname can provide compression comparable to or better than other approaches, especially for the low-bit/high-compression regimes.
We denote clustering with $b$ bits and $d$ dimensions as $b/d$  as it will assign $\frac{b}{d}$ bits in average to each weight, and the number of weight clusters is $2^{b}$. 
Especially  with multi-dim clustering such as 4/4 or 8/8 bits, our \prjname-based compression outperforms other schemes at 1 bit, while PROFIT cannot make training converge for MobileNet-v1/v2. One notable result is 64.3\% Top-1 accuracy of MobileNet-v1 with the 8/8 bit configuration (which is 1 bit-equivalent).
\prjname with 8/16 bits (effectively 0.5 bit per weight)
shows degradation from the 8/8 bit configuration, but still retains a good accuracy level. 
We also tried PROFIT+EWGS as proposed in~\citep{lee2021network}, which showed good results on
MobileNet-v1/v2 for 3  bits but failed to converge for 2 and 1 bits.

With  the overall compression ratio (or bit-per-weight) fixed, our experiments with \prjname confirm that a higher $d$ can yield a better quality training result. For the example  of MobileNet-v2, \prjname 8/16 yielded 24\% better top-1 accuracy than \prjname 4/8 although both have the same $\frac{1}{2}$ bit-per-weight, and the same trend is observed in other models. However, \prjname 8/8 failed to train ResNet50 due to the memory limitation, while \prjname 8/16 successfully trained the same model, because the   larger dimension (i.e., 8 vs 16) reduces the memory requirement of the attention matrix as discussed in Section~\ref{body:multid}.
For additional discussion, please refer to the Appendix~\ref{appendix:system}.

\begin{table}[t]
\centering
\small
\begin{threeparttable}
\renewcommand{\arraystretch}{0.95}

\begin{tabular}{cc|cccccc}
\hline
   \multicolumn{2}{c|}{ }   & ALBERT  & DistilBERT & BERT-tiny  & MobileBERT   \\  
         & Base (32 bit)                             & 90.6     & 88.2     & 78.9     &  89.6  \\  \hline \hline
    \multirow{2}{*}{{\rotatebox[origin=c]{90}{3 bit}}}
    & EWGS        & 83.3     & 87.6     &  78.3    &   87.8  \\   
    &\prjname                        & \textbf{85.1} & \textbf{88.2} &\textbf{80.0} & \textbf{89.0} \\  \hline \hline
    
     \multirow{2}{*}{{\rotatebox[origin=c]{90}{2 bit}}}
    & EWGS      & 79.6     & 85.4    &  77.9      & 81.6    \\  
    & \prjname  & \textbf{81.7} & \textbf{87.4} &  \textbf{80.0} & \textbf{83.7}\\  \hline \hline

   \multirow{3}{*}{{\rotatebox[origin=c]{90}{1 bit}}}
    & EWGS        & 62.0    & 60.9     &  74.5      &   60.2      \\ 
    & \prjname    & 79.0    & 82.8     &  \textbf{77.4}      &   69.8 \\  
    & \prjname  $4/4^\mathsection$           & \textbf{80.0}    & \textbf{84.0}    & 77.2       &  \textbf{78.3}     \\  
    

\hline

\end{tabular} 
\begin{tablenotes}
\item[] $^\mathsection$ clustering with $4$ bits and $4$ dimensions
\end{tablenotes}

\end{threeparttable}
\vspace{0.1 in}
\caption{Training performance for QNLI: \prjname-based scheme outperforms EWGS in compressing various transformed-based architectures. Also, multi-dimensional \prjname (see Section~\ref{body:multid}) largely  improved the accuracy of MobileBERT with 1 bit-per-weight target using the 4/4 configuration.}
\label{result_top1_qnli}
\vspace{-0.05 in}
\end{table}

\begin{table}[t]
\vspace{-0.15 in}
\centering
\small
\begin{threeparttable}
\setlength{\tabcolsep}{5 pt}
\renewcommand{\arraystretch}{0.95}

\begin{tabular}{cc|cccccc}
\hline 
  \multicolumn{1}{c}{ }   & Base
                                              & GOBO     & \prjname   & \prjname\\ 
      Metrics &   32bit               &  xform$^\dagger$$3$,emb$^\ddagger$$4$  &  xform$4/2^\mathsection$,emb$4$  & xform$5/2$,emb$3$\\ 
  \hline\hline   
          Top-1& 82.4         & \textbf{81.3}  & \textbf{81.3} &  \textbf{81.3} \\  
                                      Size (MB) & 255.4       & 23.9  & 21.8 &  \textbf{21.5}\\    \hline
     
\hline
\end{tabular} 
\begin{tablenotes}
\item[] $^\dagger$ the transformer layers ; $^\ddagger$ the embedding layer
\item[] $^\mathsection$ clustering with $4$ bits and $2$ dimensions
\end{tablenotes}
\end{threeparttable}
\vspace{0.1 in}
\caption{DistillBert training performance for MNLI: \prjname-based compression offered 10\% smaller model size with the same accuracy target than GOBO.}
\label{result_size_mnli}
\vspace{-0.2 in }
\end{table}

\subsection{GLUE NLP Benchmarks}
\label{result:glue}

We compared our compression by \prjname with \textbf{GOBO}~\citep{gobo} and \textbf{EWGS}~\citep{lee2021network} for BERT models on NLP tasks from  the GLUE benchmarks~\citep{wang-etal-2018-glue}, QNLI (Question-answering NLI) and MNLI (Multi NLI).
We fixed the learning rate as 1e-4 for all the experiments which worked best for EWGS, and all experiments used mini-batch size 64 per GPU (i.e., global mini-batch size of 1024) with the maximum seq-length 128.

 We compared our \prjname-based compression against EWGS~\citep{lee2021network} on the QNLI dataset, and Table~\ref{result_top1_qnli} demonstrates that \prjname offers better predictability across
all the tested models~\citep{albert,distilbert,bert_tiny,mobilebert} than EWGS. Note that the embedding layers were excluded from compression in QNLI experiments. As in ImageNet1k experiments, the 4/4 bit configuration delivers better qualities than the 1 bit configuration on all four BERT models, and especially performs well for the hard-to-compress MobileBERT.
Table~\ref{result_top1_qnli} also indicates that different transformer architectures will have different levels of accuracy degradation for a given compression target. For the example of 1 bit, MobileBERT degraded most due to many  hard-to-compress small layers, yet recovered back to a good accuracy with DKM 4/4.

When \prjname compared against GOBO~\citep{gobo} (which has outperformed the prior arts on BERT compression) on DistilBERT   with the MNLI dataset, our results in Table~\ref{result_size_mnli} clearly show that \prjname   offers a better accuracy-compression trade-off than GOBO, and also enables fine-grained balance control between an embedding layer and others: using 2.5 bits for Transformer and 3 bits for embedding is better than 2 bits for Transformer and 4 bits for embedding for DistilBERT.

\begin{table}[t]
\centering
\small
\begin{threeparttable}
\setlength{\tabcolsep}{5 pt}
\renewcommand{\arraystretch}{0.95}

\begin{tabular}{c|ccc|c|c|cccc}
\hline
          Network            & \multicolumn{3}{c|}{ResNet18}  & ResNet50 & MobileNet-v1 &MobileNet-v2 \\\hline
\multirow{2}{*}{configuration}                                 &  cv:$5/8$  & cv:$4/4$  & cv:$6/8$  & cv:$6/6$  & cv:$4/4$  & cv:$2/1$ \\  
                            &  fc:$8/4$ &  fc:$8/4$ &  fc:$6/10$ &  fc:$6/4$ &  fc:$4/2$ &  fc:$4/4$\\  
  \hline\hline   
   Inference-time    & 65.1 & 66.7 & 65.7 & 74.5 & 63.9 & 68.0\\  
   Train-time         & 66.0 & 67.0 & 66.5 & 74.7 & 65.6 & 68.8\\    
\hline
\end{tabular} 
\end{threeparttable}
 
\caption{Top-1 Accuracy with Train-time and Inference-time Weights for Fig.~\ref{result_size_imagenet1k}}
\label{dkm_train_val_discrepancy}
\end{table}

\begin{figure}[t]
	\vspace{-0.15in}
	\begin{center}
		\mbox{
			\subfigure[Top-1 accuracy]
			{\includegraphics[width=2.8in]{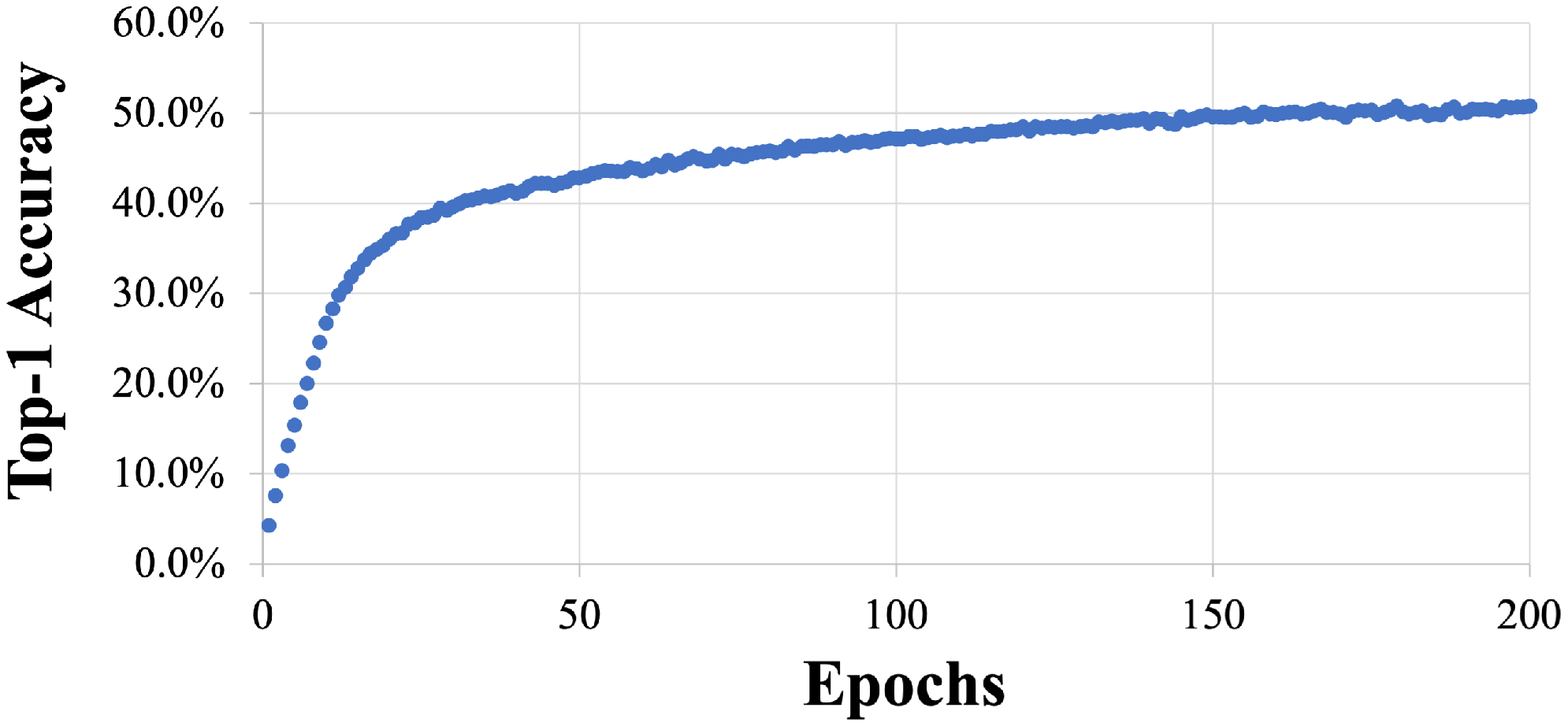}} 
			
			\subfigure[Weight distribution of the largest conv layer]
			{\includegraphics[width=2.6in]{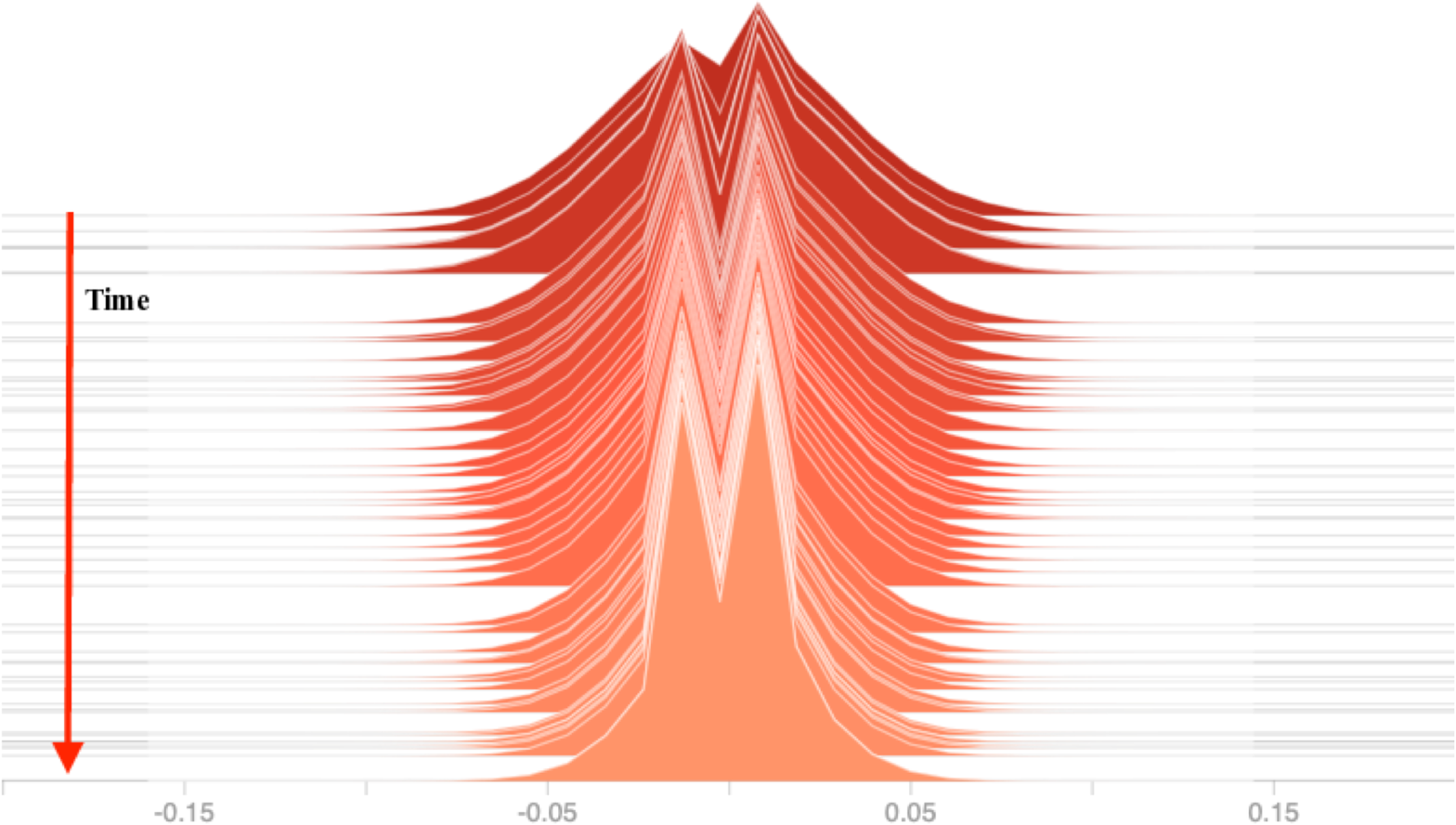}} 
		}
		\caption{MobileNet-v2 convergence with \prjname 1/1: \prjname delivers 50.8\% top-1 accuracy with 1 bit compression by gradually clustering the weights into two centroids using  the task objective only.}
		\label{mobilenet_v2_convergence}
	\end{center} 
\vspace{-0.1 in}
\end{figure}

\subsection{ \prjname-based Compression Analysis}
\label{train_inference_weight}
 Since \prjname-based compression uses attention-driven clustering as in Fig.~\ref{fig:architecture} during training but snaps the weights to the nearest centroids, there is a gap between train-time and inference-time weights which in turn leads to accuracy difference between train   and validation/test accuracies. Therefore, we   measured the Top-1 accuracies with both weights for the \prjname cases from Fig.~\ref{result_size_imagenet1k} as in Table~\ref{dkm_train_val_discrepancy}. We observed the accuracy drop is about 0.2\% - 1.7\% and gets larger with hard-to-compress DNNs.
 
 Fig.~\ref{mobilenet_v2_convergence}  shows that our compression approach can offer a smooth convergence and gradual weight-clustering based on task loss back-propagated through \prjname layers, without any custom training flow and extra regularization.  For additional details and results, please refer to Appendix~\ref{appendix:train_vs_inference_weight}. 

 



\section{Conclusion}
\label{concl}

In this work, we proposed a differentiable $k$-means clustering layer, \prjname and its application to model compression.
DNN compression powered by \prjname yields the state-of-the-art compression quality on popular computer vision and natural language models, and
especially highlights its strength in low-precision compression and quantization. The differentiable nature of \prjname
allows natural expansion to multi-dimensional $k$-means clustering, offering more than 22x model size reduction at 63.9\% top-1 accuracy for highly challenging MobileNet-v1. 

\section{Reproducibility Statement}
Our universal setup for experiments is disclosed in the first paragraph of Section~\ref{result:diffkmeans} and per-dataset-setups are also stated in the first paragraphs of
Sections~\ref{result:imagenet1k} and ~\ref{result:glue} along with key hyper-parameters in Appendix.

For Imagenet1k, we used a standard data augmentation techniques: RandomResizedCrop(224), RandomHorizontalFlip, and Normalize(mean=[0.485, 0.456, 0.406], std=[0.229, 0.224, 0.225]) as in other papers~\citep{park2020profit,lee2021network}. During evaluation, we used the following augmentations: Resize(256), CentorCrop(224), and Normalize(mean=[0.485, 0.456, 0.406], std=[0.229, 0.224, 0.225]). 
For GLUE benchmark, we used the default setup from HuggingFace.
 
\bibliography{references}
\bibliographystyle{iclr2022_conference}

\appendix
\newpage

\section{Ablation Study:   \prjname System Performance and Overhead}
\label{appendix:system}
Table~\ref{resnet_mem_runtime} shows the GPU memory utilization and per-epoch runtime for the \prjname cases from Table~\ref{result_top1_imagenet1k}. While \prjname layers have negligible impacts on training speed, the GPU memory utilization increases with more bits (i.e., more clusters) and with smaller dimensions, as   discussed in Section~\ref{body:multid}: the memory complexity of a \prjname layer with $N$ parameters and $d$ dimension is  $O(r\frac{N}{d}2^b)$ where $r$ is the number of iterations.

\begin{table}[h]
\centering
\begin{threeparttable}
\renewcommand{\arraystretch}{1.1}

\begin{tabular}{c|cc|ccccc}
\hline
    &  \multicolumn{2}{c|}{ ResNet18}  &  \multicolumn{2}{c}{ResNet50 }  \\   
    &  GPU memory   & Per-epoch   &  GPU memory & Per-epoch \\
        &  utilization (\%) &  runtime (sec)   &  utilization (\%) &  runtime (sec) \\   \hline\hline
    
           3 bit             & 23.8   & 414.1 & 55.3    & 425.0      \\   
           2 bit            & 21.7   & 401.2 & 49.2    & 429.3       \\   
           1 bit            & 20.4    & 413.3 & 45.3    & 426.3       \\  
   $4/4^\mathsection$ bit   & 22.3   & 414.4 & 51.8 &  430.4            \\  
          $4/8$ bit         & 20.7   & 423.6  & 46.7 &  410.4           \\  
          $8/8$ bit           & 51.8   & 409.4 &oom &                    \\  
          $8/16$ bit        & 35.1   & 421.9  & 78.3 &  454.6           \\  \hline
\end{tabular} 
\begin{tablenotes}
\item[]  $^\mathsection$ clustering with $4$ bits and $4$ dimensions
\end{tablenotes}

\end{threeparttable}
 
\caption{Memory and Runtime overheads from \prjname on ResNet18/50.}
\label{resnet_mem_runtime}
\end{table}

Therefore, one straightforward way to reduce GPU usage and   avoid the out-of-memory exception is to limit $r$.
In order to overcome the out-of-memory error for  ResNet50 with \prjname 8/8, we applied a constraint $r<3$ to all the \prjname layers, and could reach the top-1 accuracy of 74.0 \% with the 98.2\% GPU memory utilization. We believe the fundamental solution to this bottleneck is 
 to use a sparse representation for $\mathbf{A}$ by keeping top-$k$ centroids for each weight which is one of our future works.


\section{Ablation Study: Hyper-Parameter  $\tau$  search}

\begin{figure}[h]
	\begin{center}
		 
 \includegraphics[width=4.7in]{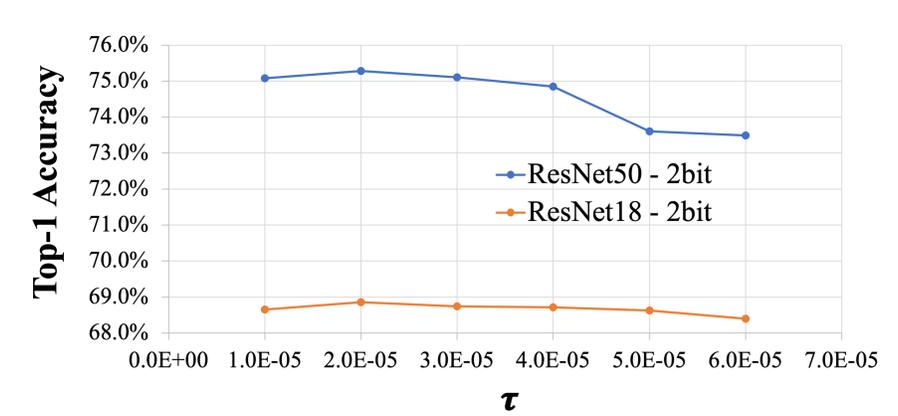}

	\caption{ResNet18/50 compression using 2 bits with varying $\tau$ values.}
		\label{resnet_tau}
	\end{center}
\end{figure}

In the current \prjname implementation, we use a global $\tau$ to control the level
 of softness in the attention matrix. The selection of $\tau$ affects the model predictive power as shown in  Fig.~\ref{resnet_tau} where  there appears to be an optimal $\tau$ for a given DNN architecture. For examples of ResNet18/80, $\tau = 2e-5$ is the best value for the 2 bit clustering.

In our experiments, we used a binary search to find out the best $\tau$ values w.r.t. the top-1 accuracy, which are listed in Table~\ref{result_tau_imagenet1k}. In general, one can observe that a complex compression task (i.e., higher compression targets, more compact networks) tends require a larger $\tau$ to provide enough flexibility or softness.

\begin{itemize}
    \item For MobileNet-v1/v2, it requires about 10x larger $\tau$ values then for ResNet18/50, because they are based on a more compact architecture and harder to compress.
    
    \item When the number of bits decreases, the compression gets harder because there are fewer centroids to utilize,  hence requiring a larger $\tau$ value.
    
    \item When the centroid dimension increases, the larger $\tau$ value is required, as the compression complexity increases (i.e., need to utilize a longer sequence). 
    
\end{itemize}

\begin{table}[h]
\centering
\begin{threeparttable}
\renewcommand{\arraystretch}{1.2}

\begin{tabular}{c|ccccccc}
\hline
     \multicolumn{1}{c|}{} & ResNet18  & ResNet50  & MobileNet-v1  & MobileNet-v2  \\   \hline\hline
    
        3 bit    & 8.0e-6      & 8.0e-6     & 5.0e-5   & 5.0e-5 \\   
         2 bit             & 2.0e-5    & 2.0e-5     & 1.0e-4         &  1.0e-4 \\   
         1 bit             & 5.0e-5    & 5.0e-5   & 3.0e-4        & 1.5e-4 \\  
         $4/4^\mathsection$ bit   & 5.0e-5 &  4.0e-5   & 1.0e-4  & 1.0e-4   \\  
         $4/8$ bit         & 5.0e-5 &  5.0e-5   & 1.0e-4  & 1.0e-4   \\  
         $8/8$ bit         & 8.0e-5 &  oom      & 1.0e-4  & 1.0e-4   \\  
         $8/16$ bit        & 1.3e-4 &  6.0e-5   & 1.2e-4  & 1.4e-4   \\  \hline

\end{tabular} 
\begin{tablenotes}
\item[]  $^\mathsection$ clustering with $4$ bits and $4$ dimensions
\end{tablenotes}

\end{threeparttable}
\vspace{0.1 in}
\caption{$\tau$ for the \prjname experiments in Table~\ref{result_top1_imagenet1k} in Section~\ref{result:diffkmeans}}
\label{result_tau_imagenet1k}
\end{table}

It could be possible to cast $\tau$ as a learnable parameter for each layer or apply some scheduling to improve the model accuracy further (as a future work), but still both approaches need a good initial point which  can be found using a binary search technique. 

For the BERT experiments with GLUE benchmarks, we used the following $\tau$ regardless of the compression level: 5.0e-5 for ALBERT, 8.0e-5 for DistilBERT, 1.5e-4 for BERT-tiny, and 4.0e-4 for MobileBERT. We found the BERT models are less sensitive to the $\tau$ than ImageNet classifiers.

\begin{figure}[t]
	\vspace{-0.15in}
	\begin{center}
		\mbox{
			\subfigure[the first conv layer in the ResBlock1]
			{\includegraphics[width=3.8in]{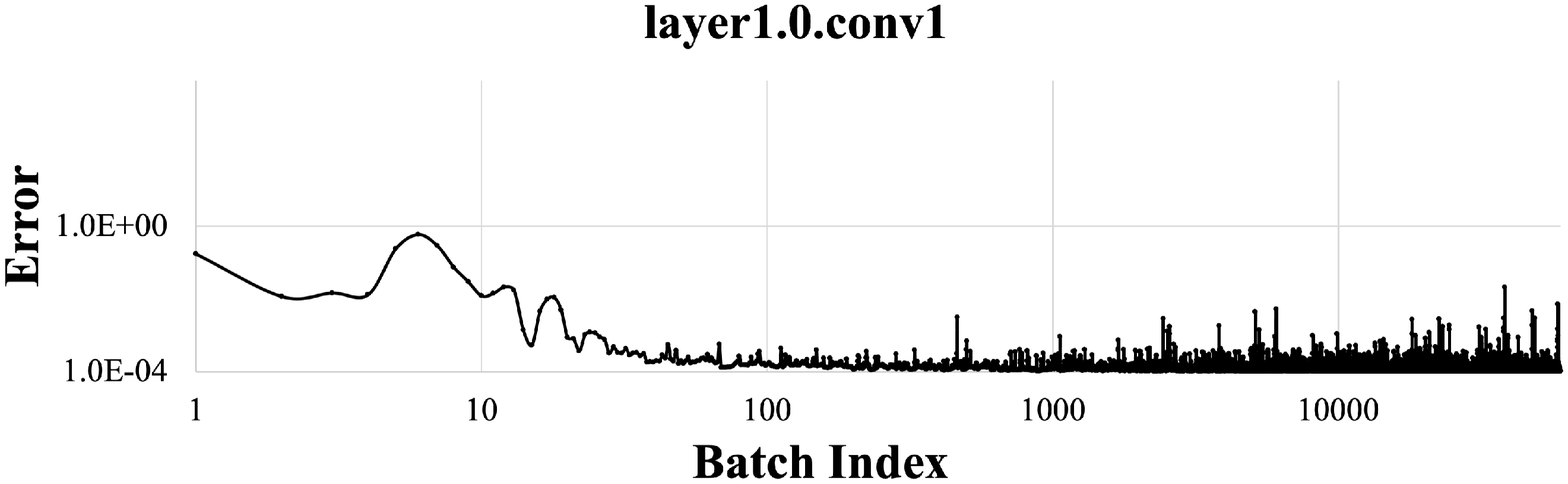}} 
		}
		\mbox{
			\subfigure[the first conv layer in the ResBlock2]
			{\includegraphics[width=3.8in]{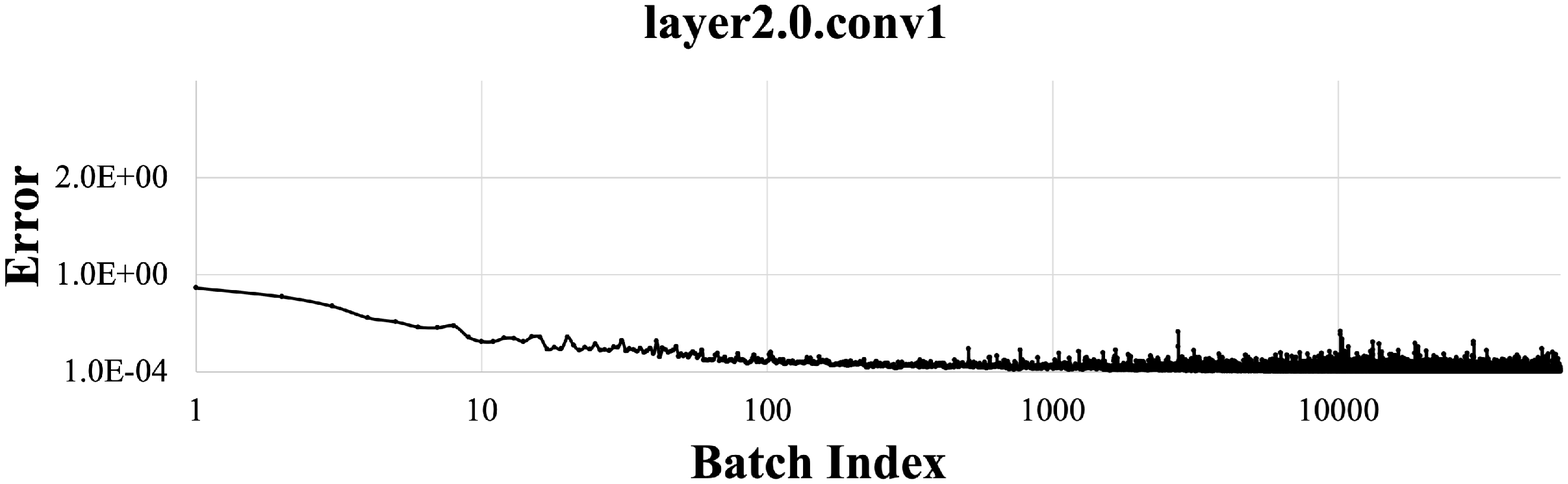}} 
		}
		\mbox{
			\subfigure[the first conv layer in the ResBlock3]
			{\includegraphics[width=3.8in]{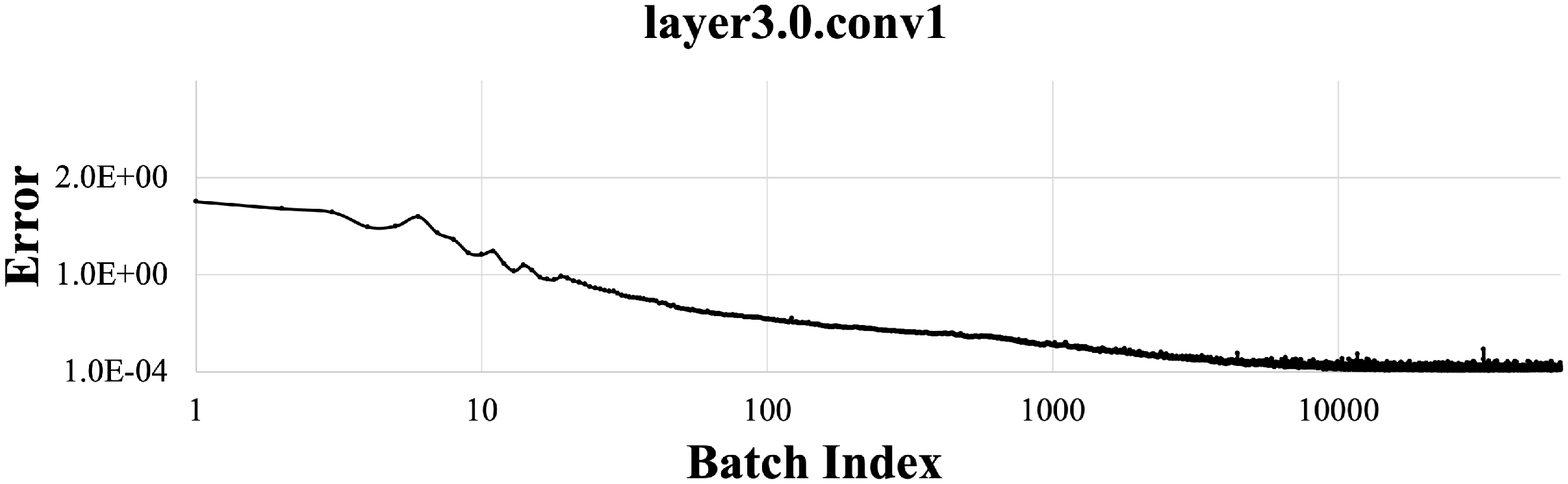}} 
		}
		\mbox{

			\subfigure[the first conv layer in the ResBlock4]
			{\includegraphics[width=3.8in]{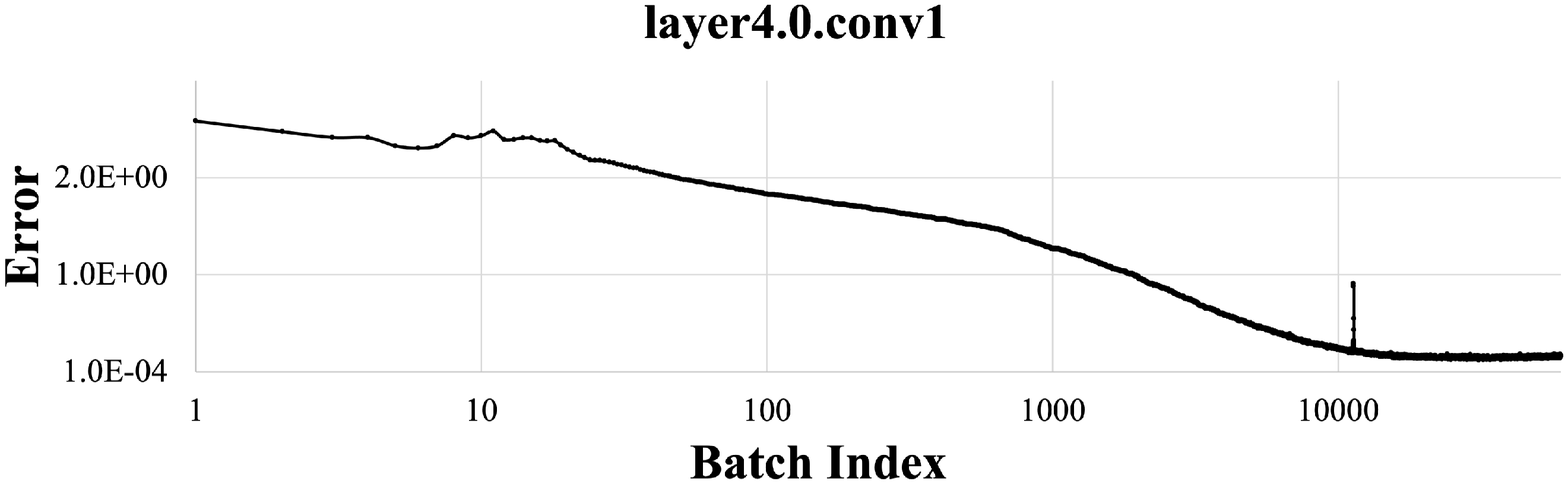}} 
		}
		\mbox{
			\subfigure[the last fc layer in ResNet18]
			{\includegraphics[width=3.8in]{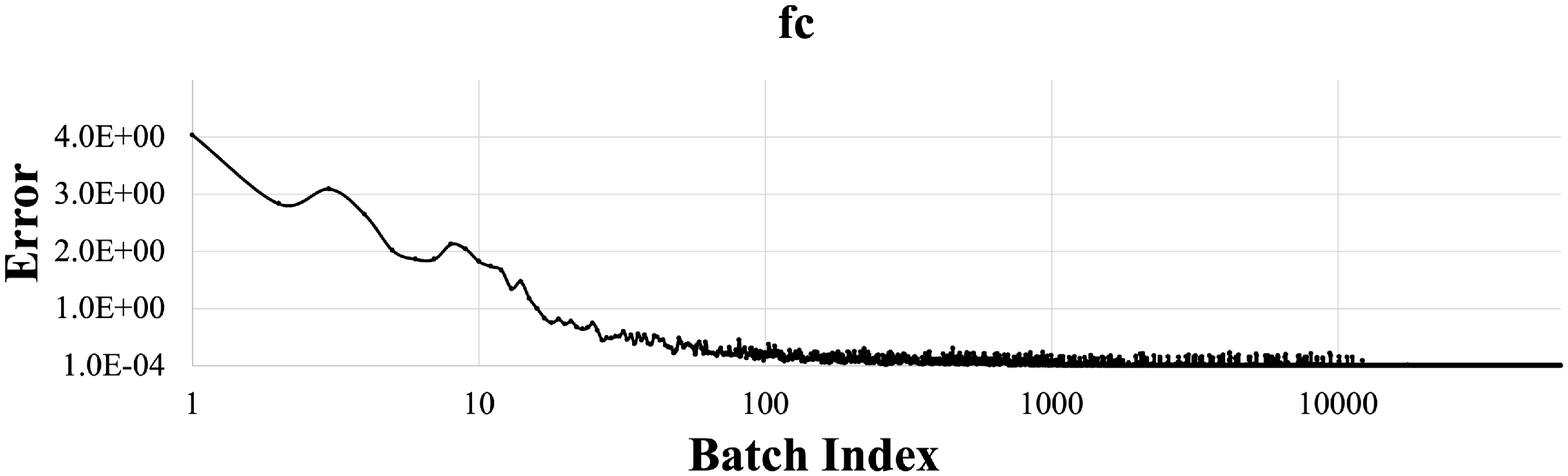}} 
		}

 	\caption{The error between train-time and inference-time weights  as a Frobenius norm for every batch for the 200 epochs are plotted, while compressing ResNet18 with \prjname cv:6/8 and fc:6/10 on ImageNet1k. After enough number of epochs, such error is reduced to the $1\text{e--}4$ level.
		}
		\label{fig:err_batch}
	\end{center} 
\end{figure}

\section{Train-time vs. Inference-time Weight Difference/Error}
\label{appendix:train_vs_inference_weight}
As discussed in Section~\ref{train_inference_weight}, \prjname-based compression requires   snap the train-time weights to the nearest centroids for accurate validation or inference, which creates minor accuracy degradation during inference. To understand the behavior better, we measured the Frobenius norm of the weight difference (i.e., $torch.norm(train\_weight - inference\_weight)$) as an error for each layer for every batch from ResNet18 with  training with \prjname cv:6/8 and fc:6/10 (from Fig.~\ref{result_size_imagenet1k}) on ImageNet1k. The  error changes of the five representative layers for over 120,000 batches or 200 epochs are plotted in Fig.~\ref{fig:err_batch} from which we can make the following observations:
\begin{itemize}
 
    \item Although every layer starts with a different level of error, the errors get smaller over the training time, and eventually at the scale of $\text{1e--}4$. 
    
    \item Aggressive compression makes a layer to begin with a high level of   error. For example, the last fc layer starts with an error of 4 (because it targets 0.6 bit per weight).
    
    \item the later layers get stabilized better than the earlier layers after enough epochs have been passed.

\end{itemize}    

Our observations are aligned with Fig.~\ref{mobilenet_v2_convergence} (b) in the sense that \prjname will encourage weights to be clustered tightly over time, decreasing the difference between train-time and inference-time weights, thus can be very effective in compressing the model and minimizing the model accuracy degradation.

\begin{figure}[!t]
	\begin{center}
		\mbox{
			\subfigure[The number of iterations across all layers combined.]
			{\includegraphics[width=4.6in]{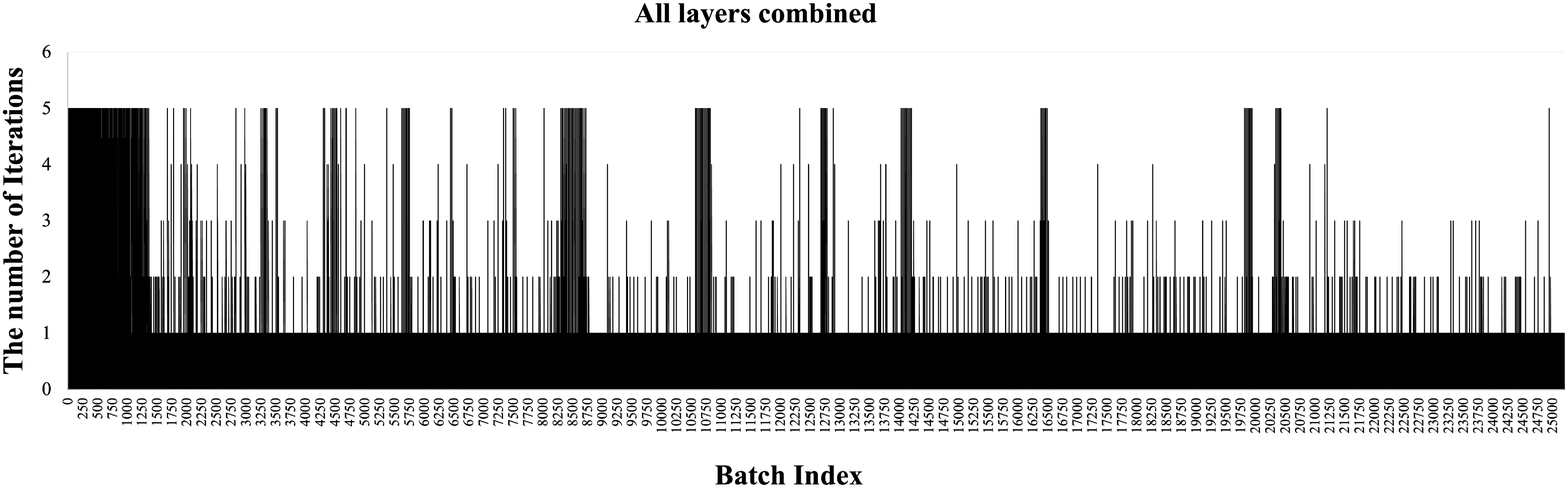}} 
			
		}

		\mbox{
			\subfigure[The 1st conv layer in the ResBlock1]
			{\includegraphics[width=2.2in]{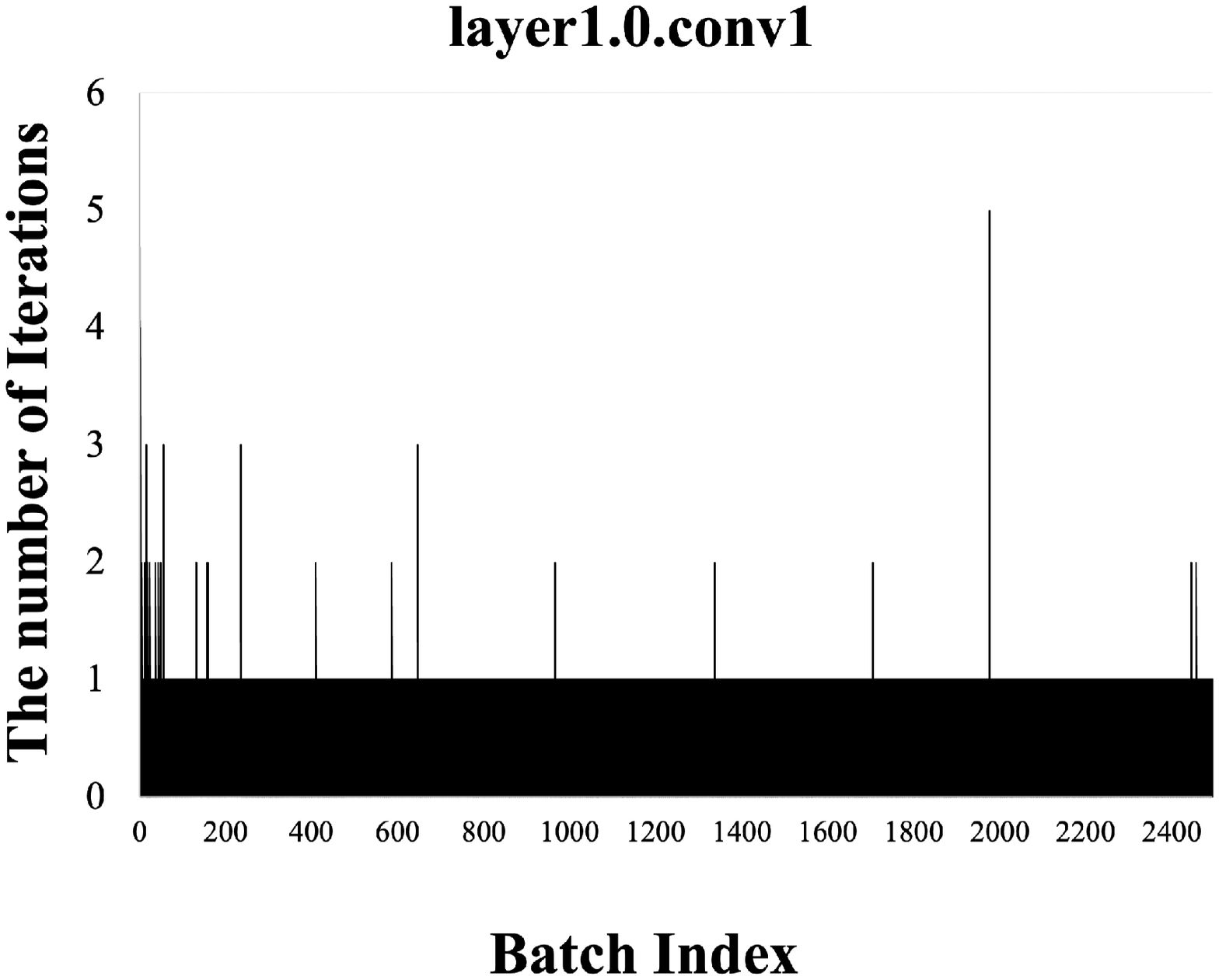}}  
			
			\subfigure[The 2nd conv layer in the ResBlock1]
			{\includegraphics[width=2.2in]{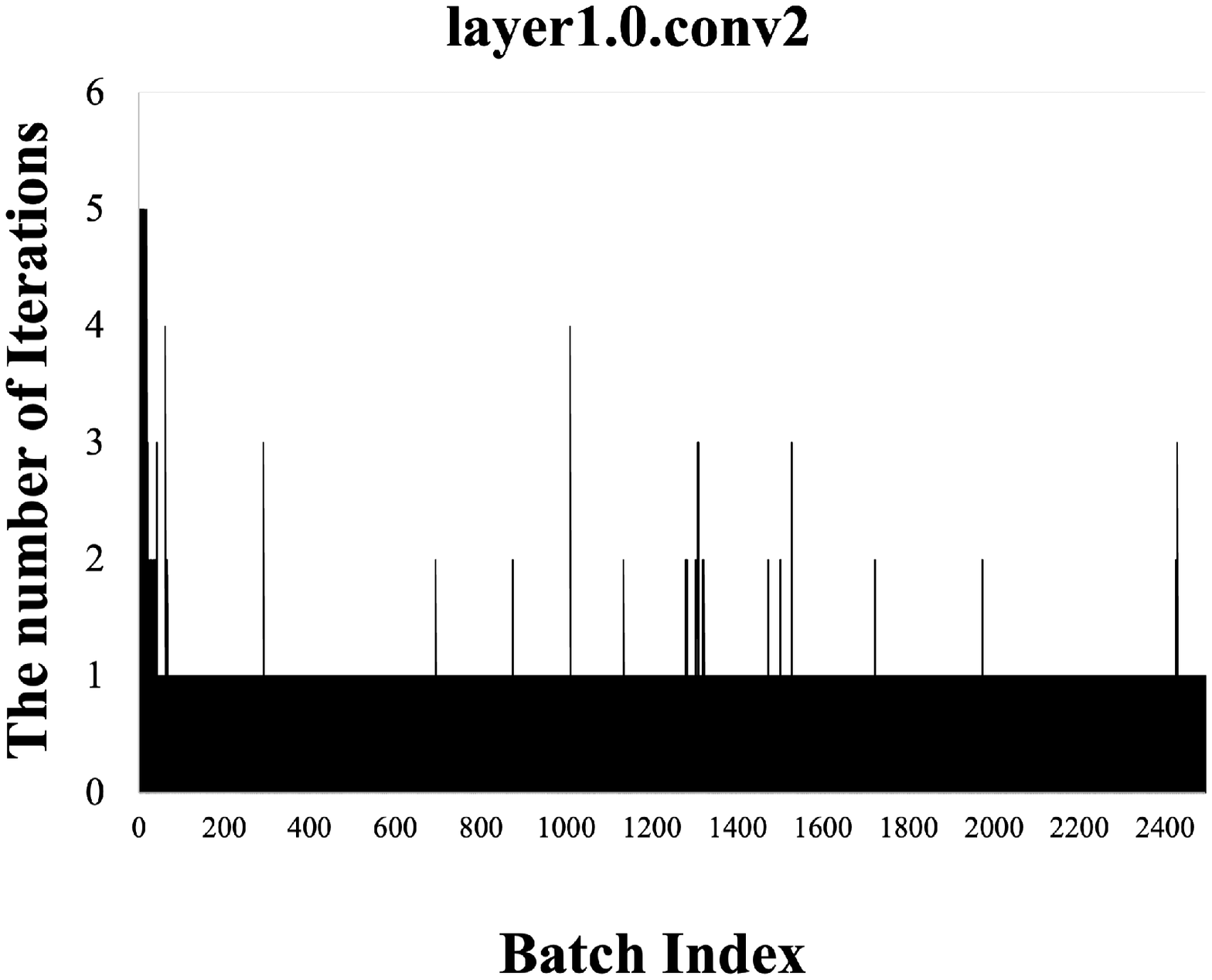}}  
		}
		
		\mbox{	
			\subfigure[The 1st conv layer in the ResBlock2]
			{\includegraphics[width=2.2in]{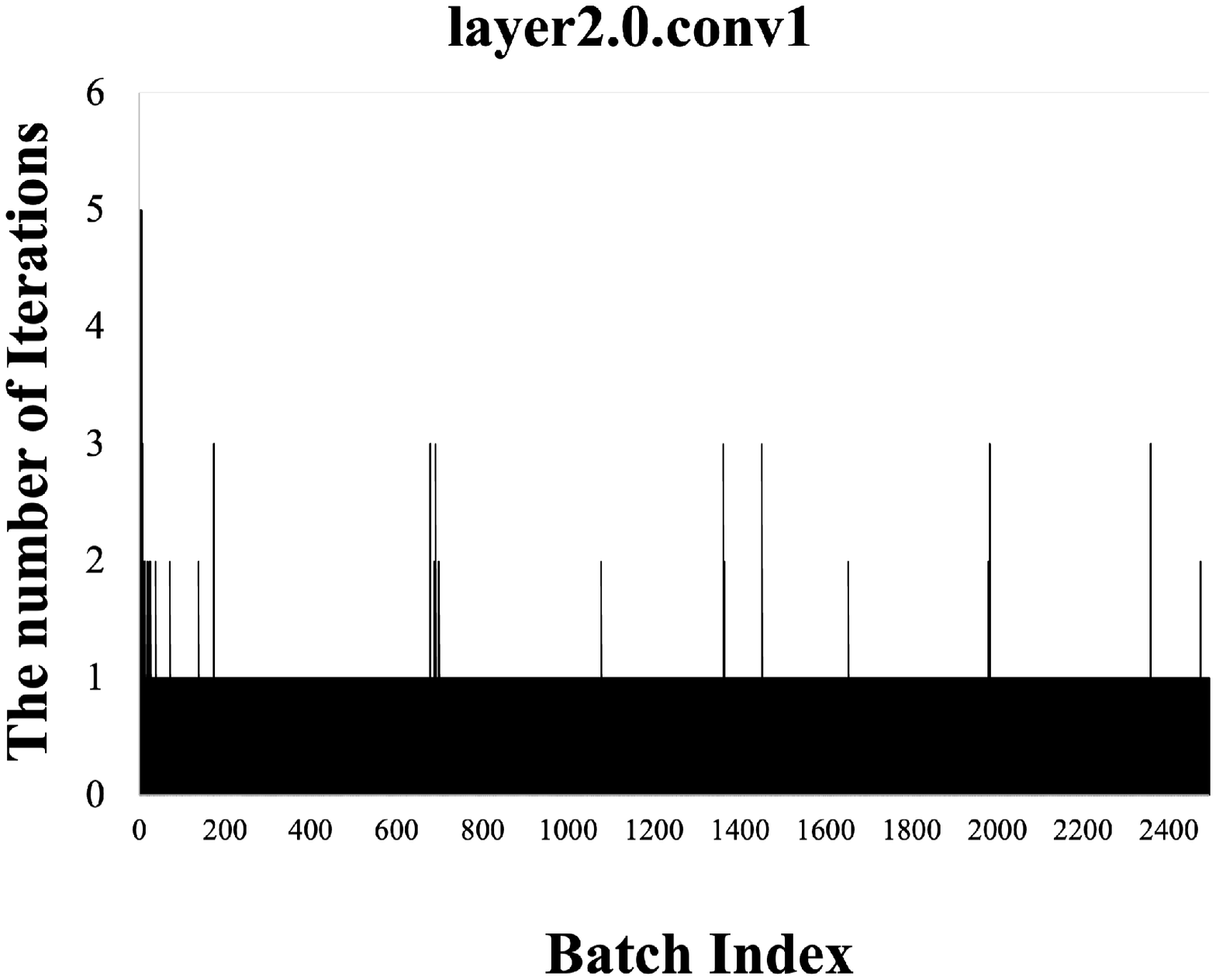}}  
 
			\subfigure[The last fc layer in ResNet18]
			{\includegraphics[width=2.2in]{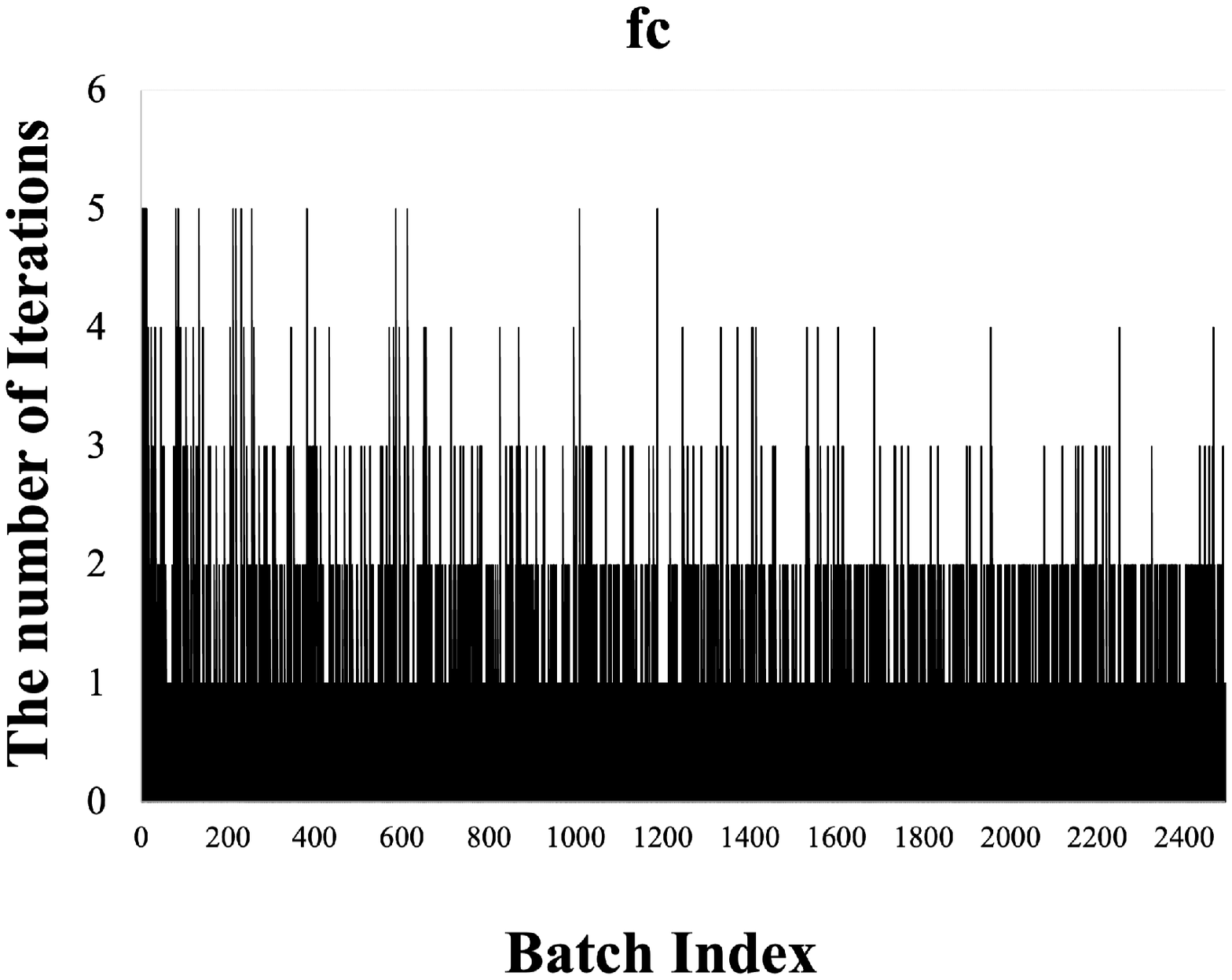}} 
		}

		
				\caption{The number of iterations in ResNet18 training with \prjname cv:6/8 and fc:6/10 on ImageNet1k.}
		\label{fig:iter_batch}
	\end{center} 
\end{figure}

\section{The number of iterations in \prjname Layer}
\label{appendix:num_iter}
In a \prjname layer, we iteratively perform $k$-means clustering using the attention mechanism until the clustering process is converged or the maximum iteration number is reached. In our experiments in Section~\ref{result:diffkmeans}, we set the maximum number of iteration as 5 to avoid the out-of-memory error. In order to understand how many   iterations are needed before each forward pass, we collected the iteration count per batch from each layer for the first eight epochs (the trend holds true for the remaining epochs) from ResNet18 with  training with \prjname cv:6/8 and fc:6/10 (from Fig.~\ref{result_size_imagenet1k}) on ImageNet1k, and plotted the graphs in Fig.~\ref{fig:iter_batch}. Our observations include the following:

\begin{itemize}
 
    \item (a) shows the number of iteration changes over the 8 epochs across all the compressed layers. As one can see, the number of iterations hits the maximum limit initially, implying that the weights are being clustered aggressively. 
    
    \item As the training continues however, the number of iterations decreases slowly with sporadic spikes, implying \prjname-based compression helps  learn good weight sharing.
    
    \item While the convolutional layers  in (b), (c), and (d) get stabilized after only a few dozens of batches, the last fc layer in (e) requires much more iterations throughout the training. We partially believe that this is because we used the \prjname 6/10 configuration which is more challenging than the \prjname 6/8 for the convolution layer.
   
\end{itemize}

\section{$\epsilon$ in \prjname}
\label{appendix:epsilon}
\begin{figure}[t]
  \centering
  \includegraphics[width=5 in]{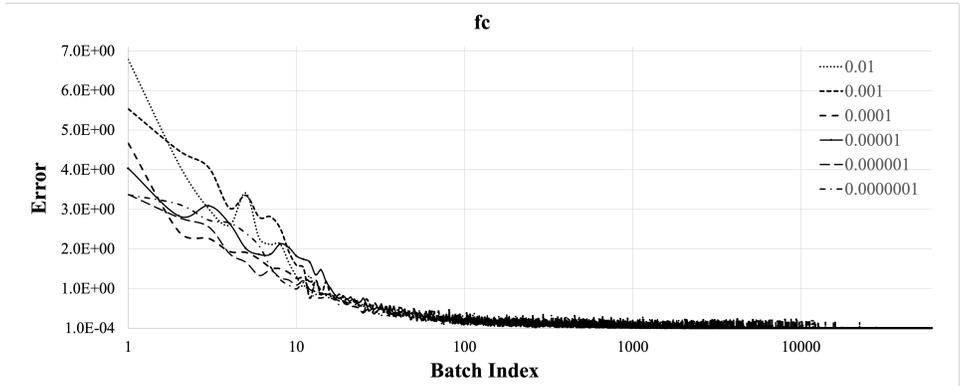}
  \caption{We plotted the train vs. inference weight error from the last fc layer of ResNet18 with \prjname cv:6/8 and fc:6/10 on ImageNet1k with varying $\epsilon$ from $1\text{e--}2$ to $1\text{e--}7$ for 200 epochs. Although the initial error is larger with bigger $\epsilon$, the final errors all converged to a similar level after a sufficient number of epochs. }
  
  \label{fig:epsilon}
\end{figure}

The $\epsilon$ in Fig.~\ref{fig:architecture} determines when to exit the iterative process. Note that the default value for the experiments in Section~\ref{result:diffkmeans} is $\text{1e--}4$
based on from sklearn.cluster.KMeans (https://scikit-learn.org/stable/modules/generated/sklearn.cluster.KMeans.html).
Therefore, we performed the sensitivity analysis by varying the $\epsilon$ from $\text{1e--}2$ to $\text{1e--}7$ for ResNet18 with \prjname cv:6/8 and fc:6/10 on ImageNet1k and increase the maximum iteration limit to 15 not to make it as a bottleneck. From the results, we found that the final Top-1 accuracies for all the cases  were in the range of 65.7 -65.9\%.
To understand why and develop insights, we plotted the train-time vs. inference-time weight difference of the fc layer  in Fig.~\ref{fig:epsilon} as similarly with Fig.~\ref{fig:err_batch} (d).  
From the Fig.~\ref{fig:epsilon}, we could make the following observations:
\begin{itemize}
 
    \item With a large $\epsilon$, there is a wider gap between training and inference weights, implying the clustering is not fully optimized. For example, the initial error  with $\epsilon=\text{1e--}2$ is about 2x larger than one with $\epsilon=\text{1e--}6$.
    
    \item However, the gap is closing over time, and after a sufficient number of epochs, the errors are all in the similar range of $\text{1e--}4$ regardless of the $\epsilon$ value.
    
    \item Larger $\epsilon$ values make \prjname layers iterate fewer, decreasing the peak memory consumption. For example, the iteration count was 1 throughout training when $\epsilon=\text{1e--}2$.
     
\end{itemize}    
 Therefore, apparently, as long as the training with \prjname layers can run long enough, the selection of $\epsilon$ might not affect the final result. 
We believe this is because \prjname ensures the clustering continuity by resuming from the last known centroid (i.e., from the previous batch). 
In case that the planned training time is short, a smaller $\epsilon$ value would be preferred but at the cost of larger memory requirement.

\section{Gumbel-softmax for soft assignment}
\label{appendix:gbs_hard}
  Gumbel-Softmax distribution is a continuous distribution that approximates samples from a categorical distribution and also works with back-propagation. Therefore, it is feasible to use the drawing from Gumbel-softmax as soft assignment to generate attention.
Therefore, we experimented  Gumbel-Softmax-based \prjname (\textbf{GBS}) along with the hard assignment scheme (\textbf{Hard}) with ResNet18 on ImageNet1k. 
We ran clustering iteratively for both GBS and Hard, and such iteration helps GBS reduce the variance~\citep{lr_net}.

\begin{figure}[!t]
 
    \centering
  \includegraphics[width=5 in]{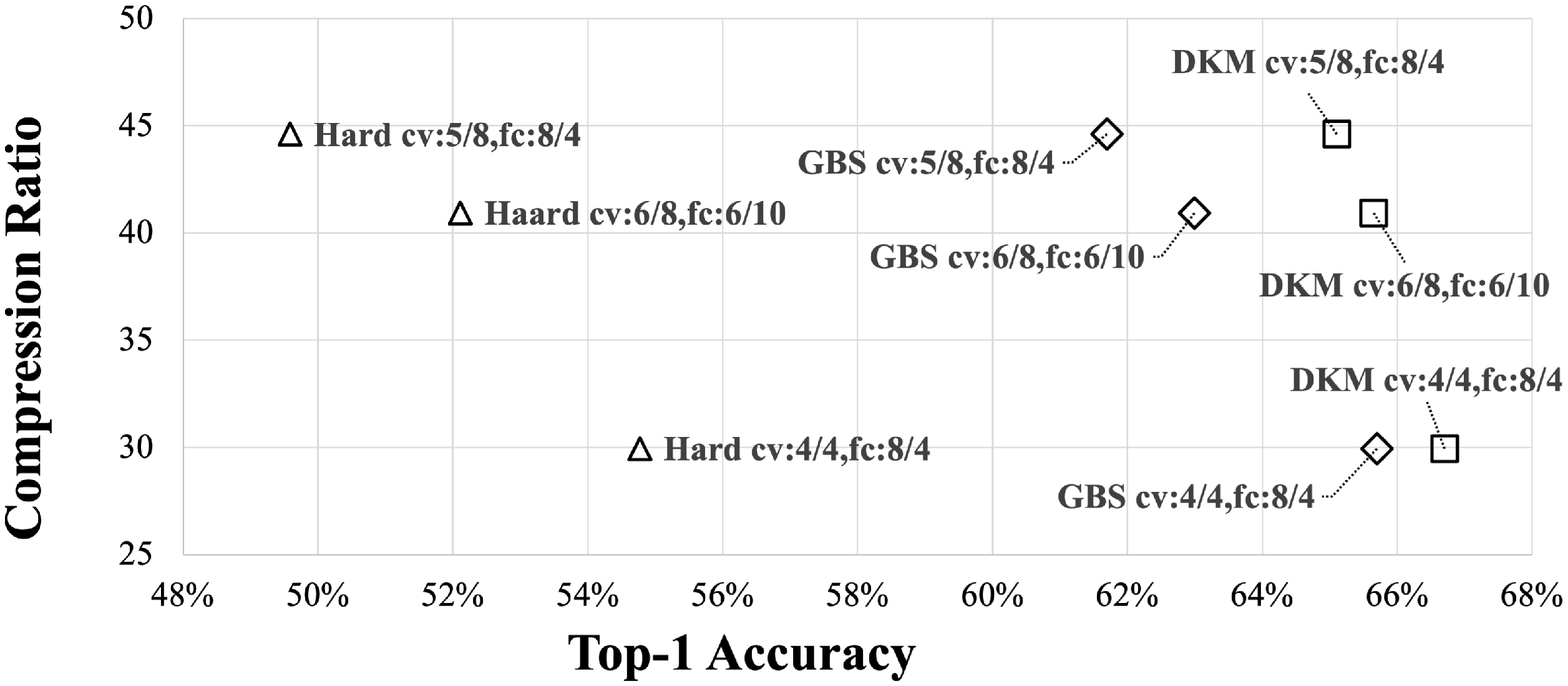}
\vspace{0.1 in}
\begin{threeparttable}
\setlength{\tabcolsep}{5 pt}
\renewcommand{\arraystretch}{0.9}

\vspace{0.1 in}
\begin{tabular}{c|cccc|c|cccc}
\hline
 {configuration}   &   cv:$4/4$,fc:$8/4$   & cv:$6/8$,fc:$6/10$  & cv:$5/8$,fc:$8/4$  \\  
  \hline\hline   
  \prjname &   66.7  &65.7& 65.1\\  
  GBS   &    65.7 &62.9 & 61.7\\  
  Hard  &    54.9 & 52.1 &49.4\\  
 
\hline
\end{tabular} 
\end{threeparttable}
\caption{ When \prjname is compared with another soft assignment method, \textbf{GBS} based on Gumbel-softmax and a hard assignment method, \textbf{Hard} with ResNet18 on ImageNet1k, \prjname outperforms both. Soft assignment techniques outperform hard by a wide margin, and \prjname is superior to GBS.}
\label{dkm_gsm_hard}
  \end{figure}

We applied the same compression configurations from Fig.~\ref{result_size_imagenet1k} and kept all hyper-parameters and training flow intact.
The comparison results are in Fig.~\ref{dkm_gsm_hard} which shows that Hard is much worse than both GBS and \prjname on all the cases with over 10\% drop in Top-1 accuracy, proving that soft assignment is a superior way of clustering weights for model compression.
Although GBS outperforms Hard, GBS is still worse than \prjname on all cases, and the degradation can be as significant as 3.4\% drop in Top-1 accuracy for the most aggressive compression target.
 
\section{Relation to Expectation-Maximization (EM) and theoretical interpretation} 
\label{appendix:em}

Special sub-case of \prjname where gradients are not propagated can be related to a standard EM formulation. Here is the formulation-level correspondence: Suppose $$p(x) = \sum_{i=1}^{K} \frac{1}{K} \mathcal{N}(x|c_i, \sigma^2=\tau/2)$$ is the Gaussian mixture model over cluster centers. Referring to Fig.~\ref{fig:architecture}:
\begin{itemize}
\item Maximizing log likelihood of weights $$\ln P(W|C) = \sum_{i=1}^{N} \ln \big\{ {\sum_{j=1}^{K} \frac{1}{K} \mathcal{N}(w_i|c_j,\sigma^2=\tau/2})\big\}$$ using the EM algorithm is equivalent to \prjname for the case of $d_{i,j}= -(w_i-c_j)^2$.
\item The attention matrix $\mathbf{A}$ where $a_{i,j} = \frac{exp(\frac{d_{i,j}}{\tau})}{\sum_{k}exp(\frac{d_{i,k}}{\tau})}$ is equivalent to the responsibilities calculated in the E step: $r_{i,j} = \frac{\frac{1}{K}\mathcal{N}(w_i|c_j,\sigma^2=\tau/2)}{\sum_{k=1}^K \frac{1}{K}\mathcal{N}(w_i|c_k,\sigma^2=\tau/2)} = a_{i,j}$ 
\item Updating $\mathbf{C}$  using $c_j = \frac{\sum_{i} {a}_{i,j} {w}_i}{\sum_{i} {a}_{i,j}}$ is equivalent to M step of EM algorithm. Notice that variance is fixed, therefore M step in EM is only updating cluster centers as DKM does.

\end{itemize}

However, unlike EM where finding $\mathbf{C}$ every M step is the objective, \prjname   focuses on
generating a representative $\mathbf{\tilde W}$ for the train-time compression for DNN.

Even though there is formulation-level similarity between \prjname and EM, the way both are optimized is significantly different.
While EM iteratively optimizes a specific likelihood function for a set of \textbf{fixed} observations, 
\prjname needs to adjust (i.e., optimize) the observations (which are weights) without leading to a trivial solution such as all observations collapsing to a certain point.
Hence, \prjname can neither assume any statistical distribution nor optimize a specific likelihood function (i.e., the observations are dynamically changing).
    Therefore, \prjname uses a simple softmax and rides on the back-propagation to fine-tune the observations w.r.t. the task
    loss function after unrolling multiple attention updates. When we propagate gradients, then this will turn into a stochastic non-convex joint optimization where we simultaneously optimize observations and centroids for the task loss function, 
    which is shown to offer better accuracy vs. compression trade-offs according to our experiments.


\end{document}